\documentclass{article} 
\usepackage[T1]{fontenc}
\usepackage{iclr2027_conference,times}

\usepackage{amsmath,amsfonts,bm}

\def\eqref#1{equation~\ref{#1}}

\def\1{\bm{1}}

\DeclareMathAlphabet{\mathsfit}{\encodingdefault}{\sfdefault}{m}{sl}
\SetMathAlphabet{\mathsfit}{bold}{\encodingdefault}{\sfdefault}{bx}{n}

\usepackage{hyperref}
\usepackage{url}
\usepackage{booktabs}
\usepackage{graphicx}
\usepackage{multirow}
\usepackage{xcolor}
\usepackage{array,colortbl}
\definecolor{CMRink}{HTML}{222B33}
\definecolor{CMRaccent}{HTML}{365B73}
\definecolor{CMRline}{HTML}{BCC4CA}
\definecolor{CMRmuted}{HTML}{A1ABB4}
\definecolor{CMRnote}{HTML}{57626B}

\newcommand{\CMRTableSetup}{%
  \fontsize{8.5}{10.5}\selectfont
  \color{CMRink}%
  \setlength{\tabcolsep}{0pt}%
  \renewcommand{\arraystretch}{1.12}%
  \arrayrulecolor{CMRink}%
}

\newcommand{\CMRGroup}[1]{\textcolor{CMRaccent}{\bfseries #1}}
\newcommand{\CMRMetric}[1]{\textcolor{CMRnote}{\fontsize{7.5}{9}\selectfont #1}}

\usepackage{amsmath}
\usepackage{subcaption}
\usepackage{tikz}
\usepackage{float}
\usepackage{wrapfig}
\usetikzlibrary{arrows.meta,positioning}

\newcommand{\stageone}{Stage-1}
\newcommand{\stagetwo}{Stage-2}
\newcommand{\nQAone}{128{,}915}
\newcommand{\nQAtwo}{42{,}799}

\title{Learning Where to Look: Anatomical Grounding and
Guided Attention for Cardiac MRI Vision--Language Models}
\author{%
Bangwei Guo\textsuperscript{1,2}\thanks{This work was carried out during the internship of the authors at United Imaging Intelligence, Boston, MA}\hspace{0.65em}
Xiao Chen\textsuperscript{2}\hspace{0.65em}
Boris Mailhe\textsuperscript{2}\hspace{0.65em}
Jia Yao\textsuperscript{2,3}\footnotemark[1]\\[1pt]
\bfseries Yiqing Wang\textsuperscript{2,4}\footnotemark[1]\hspace{0.65em}
Ankush Mukherjee\textsuperscript{2}\hspace{0.65em}
Yikang Liu\textsuperscript{2}\\[1pt]
\bfseries Zheyuan Zhang\textsuperscript{2}\hspace{0.65em}
Hang Yu\textsuperscript{2}\hspace{0.65em}
Terrence Chen\textsuperscript{2}\hspace{0.65em}
Shanhui Sun\textsuperscript{2}\thanks{Corresponding author: \href{mailto:shanhui.sun@uii-ai.com}{\texttt{shanhui.sun@uii-ai.com}}.}\\[6pt]
{\small\normalfont\textsuperscript{1}Rutgers University, NJ, USA}\\[1.5pt]
{\small\normalfont\textsuperscript{2}United Imaging Intelligence, Boston, MA}\\[1.5pt]
{\small\normalfont\textsuperscript{3}University of Texas Southwestern Medical Center, TX, USA}\\[1.5pt]
{\small\normalfont\textsuperscript{4}Duke University, NC, USA}
}
\hypersetup{
  pdftitle={Learning Where to Look: Anatomical Grounding and Guided Attention for Cardiac MRI Vision--Language Models},
  pdfauthor={Bangwei Guo, Xiao Chen, Boris Mailhe, Jia Yao, Yiqing Wang, Ankush Mukherjee, Yikang Liu, Zheyuan Zhang, Hang Yu, Terrence Chen, Shanhui Sun}
}
\begin{document}
\maketitle
\raggedbottom

\begin{abstract}
Cardiac magnetic resonance imaging (CMR) enables assessment
of cardiac anatomy, ventricular function, and myocardial
tissue characteristics.
Clinicians interpret these images by identifying cardiac
structures and focusing on the regions relevant to each
clinical question, motivating anatomically guided vision--language models (VLMs). Yet CMR-specific supervision for anatomical localisation
and clinical question answering remains limited. To address this gap, we investigate fine-grained CMR visual question answering through anatomical grounding
and guided attention. We construct \nQAone{} anatomical-grounding and
\nQAtwo{} clinical QA pairs across short-axis cine,
late gadolinium enhancement, and long-axis cine.
These datasets support anatomical recognition, localisation,
and clinical assessment without requiring paired reports
for individual training images. To help the model \emph{learn where to look}, we introduce
Cardiac Anatomy-Routed Attention (CARA), which selects
predicted anatomical priors according to the question
and guides decoder attention with learned task-specific
strengths.
Combining anatomical grounding pretraining with CARA
yields our model, CARA-VL.
Experiments demonstrate CARA-VL's strengths in clinical
assessment and regional localisation across CMR imaging
settings, with promising generalization to an external
clinical cohort.
Together, our data and method provide a practical framework
for studying and advancing cardiac visual understanding
in VLMs. We will release the QA data derived from public datasets
upon publication.
\end{abstract}


\section{Introduction}
\label{sec:intro}

Cardiac magnetic resonance imaging (CMR) enables comprehensive assessment of
cardiac anatomy, ventricular function, and myocardial tissue
characteristics~\citep{schulz2020standardized}. Clinicians interpret CMR by first identifying the relevant cardiac
structures and then assessing their morphology, motion, and tissue
characteristics for the diagnosis. For example, hypertrophy is assessed from myocardial wall thickness, ventricular function from changes across cardiac phases, and myocardial injury from the presence, extent, and transmurality of scar on late gadolinium enhancement (LGE)~\citep{kim2000use,schulz2020standardized}. However, reliable CMR interpretation requires specialist
expertise and often time-consuming quantitative analysis.
This motivates cardiac visual question answering (VQA),
in which AI models answer clinically relevant questions
about cardiac anatomy, function, and tissue characteristics
from CMR images.

Recent medical vision--language models (VLMs) have shown strong capability
in medical image understanding and VQA through
large-scale multimodal pretraining and medical instruction tuning
\citep{li2023llava,moor2023med,
chen2024huatuogptvision,sellergren2026medgemma}. Yet their visual training is largely based on broad online biomedical
image--text corpora and commonly used medical imaging datasets, with
comparatively limited supervision on cardiac MRI. This specialization gap
can lead to poor generalization on fine-grained CMR tasks~\citep{o2026marcus}. Motivated by this gap, cardiac-specific vision--language models have recently
begun to emerge, using CMR--report pairs, cardiac-specific pretraining, or
modality-specialized experts to improve domain adaptation
\citep{nakashima2026contrastive,shad2026generalizable,o2026marcus, qu2026baai}.
Despite these advances, accessible training data and
explicit anatomical grounding remain important challenges.
Many cardiac VLMs rely on institutional or non-public
CMR cohorts, limiting access to their training data
and reproducibility.
Moreover, clinical answer supervision alone does not
explicitly teach models to localise cardiac structures
or guide attention towards the anatomy relevant to
each question.


To address these limitations, we introduce an anatomically
grounded vision--language model for fine-grained CMR VQA,
connecting anatomical recognition and localisation with
clinical assessment. We construct complementary
anatomical-grounding and clinical VQA datasets from public
and in-house CMR resources. Expert segmentation annotations
provide \nQAone{} grounding QA pairs, while quality-controlled
image-derived measurements yield \nQAtwo{} clinical QA pairs
across 14 tasks informed by clinical report analysis.
These datasets support learning about cardiac structure,
function, and myocardial scar without requiring a paired
report for each training image. Clinicians direct their attention to different cardiac
structures depending on the clinical question, such as
the myocardium for hypertrophy and the ventricular cavities
for size and function \citep{schulz2020standardized}.
Inspired by this practice, we introduce
\textbf{Cardiac Anatomy-Routed Attention (CARA)}
to help the model \emph{learn where to look}.
CARA predicts soft anatomical occupancy maps from visual
features and selects the relevant map for each clinical
question. The resulting prior guides decoder attention
over visual tokens with a learned task-specific strength
(Figure~\ref{fig:two-stage-cara}).
We refer to the resulting model, trained with anatomical
grounding and CARA, as CARA-VL.
Our main contributions are summarised as follows:

\begin{itemize}
\item We systematically investigate cardiac visual
understanding in vision--language models through
fine-grained CMR VQA, examining how anatomical recognition
and localisation support clinical assessment of cardiac
structure, function, and myocardial scar.

\item We construct \nQAone{} anatomical-grounding and
\nQAtwo{} clinical QA pairs across SAX cine, LGE, and
LAX cine. Report-guided task selection, segmentation
annotations, and quantitative measurements provide
complementary supervision without requiring paired
reports for individual training images.

\item We develop CARA-VL, combining anatomical grounding
pretraining with CARA to learn where to look for clinical
question answering. CARA routes predicted anatomical
priors according to the question and guides decoder
attention with learned task-specific strengths.

\item Experiments across 14 clinical tasks show that
CARA-VL outperforms the evaluated baselines on all
internal tasks and demonstrates promising external
generalization. Ablations support the complementary
benefits of anatomical grounding pretraining and CARA.
\end{itemize}

\section{Clinically grounded QA dataset construction}
\label{sec:data}

\begin{figure}[t]
\centering
\includegraphics[width=\linewidth]{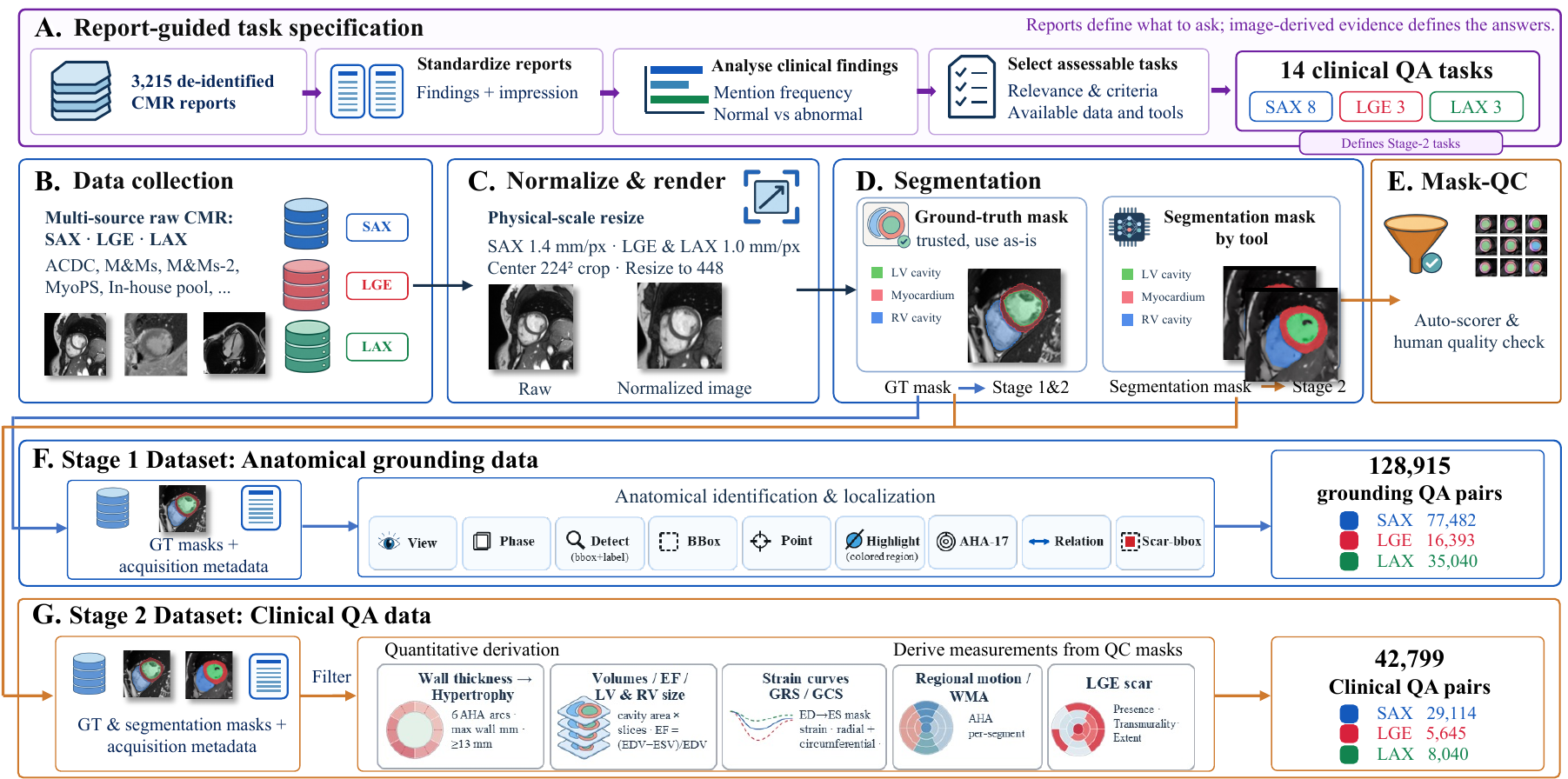}
\caption{CMR VQA dataset construction pipeline.
(A) Clinical report analysis guides the selection of
14 assessable clinical QA tasks.
(B--E) Multi-source CMR images are standardised and
paired with expert or quality-controlled automatic masks.
(F) Expert masks and acquisition metadata generate
nine types of anatomical-grounding QA for identifying
and localising cardiac structures.
(G) Image-derived measurements and task-specific decision
rules generate clinical QA on cardiac structure,
function, and myocardial scar.}
\label{fig:datapipe}
\end{figure}

We construct complementary anatomical grounding and
clinical QA datasets (Figure~\ref{fig:datapipe}).
Below, we describe report-guided task selection,
image collection and standardisation, and the construction
of anatomical grounding and clinical QA pairs.

\subsection{Report-guided clinical QA task specification}
\label{sec:questions}

We define clinical QA tasks through report standardisation,
finding-level frequency analysis, and clinically informed
selection. The corpus comprises 3{,}215 de-identified CMR
reports from three sources. Preprocessing standardises
the findings and impression sections and groups synonymous
clinical expressions. We quantify the report-level frequency
of each clinical finding, distinguishing normal from abnormal
mentions to identify candidate tasks. The final selection
considers clinical relevance, the availability of explicit
measurement or grading criteria, and whether the available
data and analysis tools support reliable reference answers.
The resulting 14 clinical QA tasks SAX
cine, LGE, and LAX cine. Eight SAX cine tasks assess hypertrophy,
its segmental distribution, left ventricular (LV) size,
right ventricular (RV) size, ejection fraction (EF),
regional wall motion, global radial strain (GRS), and
global circumferential strain (GCS). Three LGE tasks assess
scar presence, transmurality, and extent, while three
LAX cine tasks assess EF, end-diastolic volume (EDV),
and global longitudinal strain (GLS). Each task has a predefined answer set and explicit
assessment criteria, as summarised in Section~\ref{sec:stage2data} and more details are provided in Appendix~\ref{app:datadetails}.

\subsection{Image data collection and standardization}
\label{sec:sources}

We combine public datasets and in-house CMR cohorts
to construct large-scale anatomical grounding and
clinical QA datasets for pretraining, clinical adaptation,
and evaluation. The collection covers SAX cine, LGE,
and LAX cine. Public sources include
ACDC~\citep{bernard2018deep},
M\&Ms, M\&Ms-2~\citep{campello2021multi, martin2023deep},
MyoPS-related cohorts~\citep{zhuang2018multivariate,
qiu2023myops,ding2023aligning,ding2025cinemyops},
EMIDEC~\citep{lalande2020emidec},
CMR-MULTI~\citep{qu2026baai}, and the Kaggle Data
Science Bowl Challenge~\citep{kaggle2015cardiac}.
The public datasets used here do not provide paired
clinical reports. Our in-house data comprise an image-only
collection (In-house A) used for training, and an image--report
cohort (In-house B) reserved exclusively for external evaluation. To standardize the physical scale of images within each
imaging modality, we resample SAX cine to 1.4\,mm and
LGE and LAX cine to 1.0\,mm in-plane resolution using
acquisition-specific pixel spacing. Images and available masks are center-cropped to
$224 \times 224$ pixels. The images are then upsampled to $448 \times 448$ for model input, with
grounding coordinates scaled accordingly. Available ground-truth segmentation masks are used
directly for data generation. For images without
reference masks, we obtain segmentations from a deployed
internal CMR segmentation pipeline and manually review their quality
before clinical QA construction.
Details are provided in Appendix~\ref{app:datadetails}.

\subsection{Stage-1: anatomical grounding dataset}
\label{sec:stage1data}

Before clinical QA adaptation, we first pretrain the VLM to recognize and
localize cardiac anatomy as a structural foundation for subsequent reasoning,
consistent with recent evidence on anatomical grounding in medical VLMs
\citep{gu2026anatomy,zhang2026medground}. Accordingly, from the multi-source CMR pool described in
Section~\ref{sec:sources}, we select expert-annotated cases for anatomical
grounding QA, covering 878 SAX patients, 440 LAX series, and 2,241 LGE
slices. We restrict Stage-1 to ground-truth segmentation masks, allowing
all grounding targets to be derived deterministically from the annotations. The resulting nine QA task types cover imaging context, anatomical
identification and localisation, and cardiac-specific spatial
reasoning (Table~\ref{tab:stage1tasks}). Specifically, view and phase
establish imaging context; detection, bounding-box, point, highlighted-region,
and scar bounding-box tasks provide structure- and region-level grounding;
and segment and relation tasks supervise identification of American
Heart Association (AHA) myocardial segments (e.g., anteroseptal and
inferolateral segments) and LV--RV spatial relationships, respectively. Together, these tasks teach the model what structures
are present, where they are located, and how they relate spatially. The final Stage-1 dataset contains 128{,}915 QA pairs over 45{,}773 images,
including 77{,}482 SAX cine, 35{,}040 LAX cine, and 16{,}393 LGE pairs.
Further details are provided in Appendix~\ref{app:datadetails}.

\subsection{Stage-2: clinical QA dataset}
\label{sec:stage2data}

Stage-2 extends anatomical grounding to clinical assessment through
the 14 tasks defined in Section~\ref{sec:questions}. We construct
reference answers from anatomical measurements, available reference
volumes, and feature-tracking outputs, enabling clinical QA generation
from images without paired reports. These quantities capture wall
thickness, chamber size, ventricular function and deformation, and
myocardial scar. Task-specific decision rules map them to binary
findings, severity grades, or regional labels. All masks used for answer generation undergo manual quality control, with low-quality or unreliable masks excluded before deriving clinical measurements and reference answers. Each accepted label is paired with a question template specifying
the imaging context and a canonical answer. Depending on the task,
the input contains a single image or an ordered end-diastolic (ED)
and end-systolic (ES) pair. The model receives
the images and question, without mask overlays or reference
measurements. The resulting dataset contains \nQAtwo{} QA pairs:
29{,}114 from SAX cine, 5{,}645 from LGE, and 8{,}040 from LAX cine.
Patient-level exclusions are shared across tasks, views, and both
training stages, with the image--report cohort reserved for external
evaluation. Details of measurement construction, grading, review,
and sampling are provided in Appendix~\ref{app:dd-stage2}; split
definitions and evaluation references are given in
Appendix~\ref{app:dd-splits}.

\section{CARA: Cardiac Anatomy-Routed Attention}
\label{sec:method}

\begin{figure*}[t]
\centering
\includegraphics[width=\textwidth]{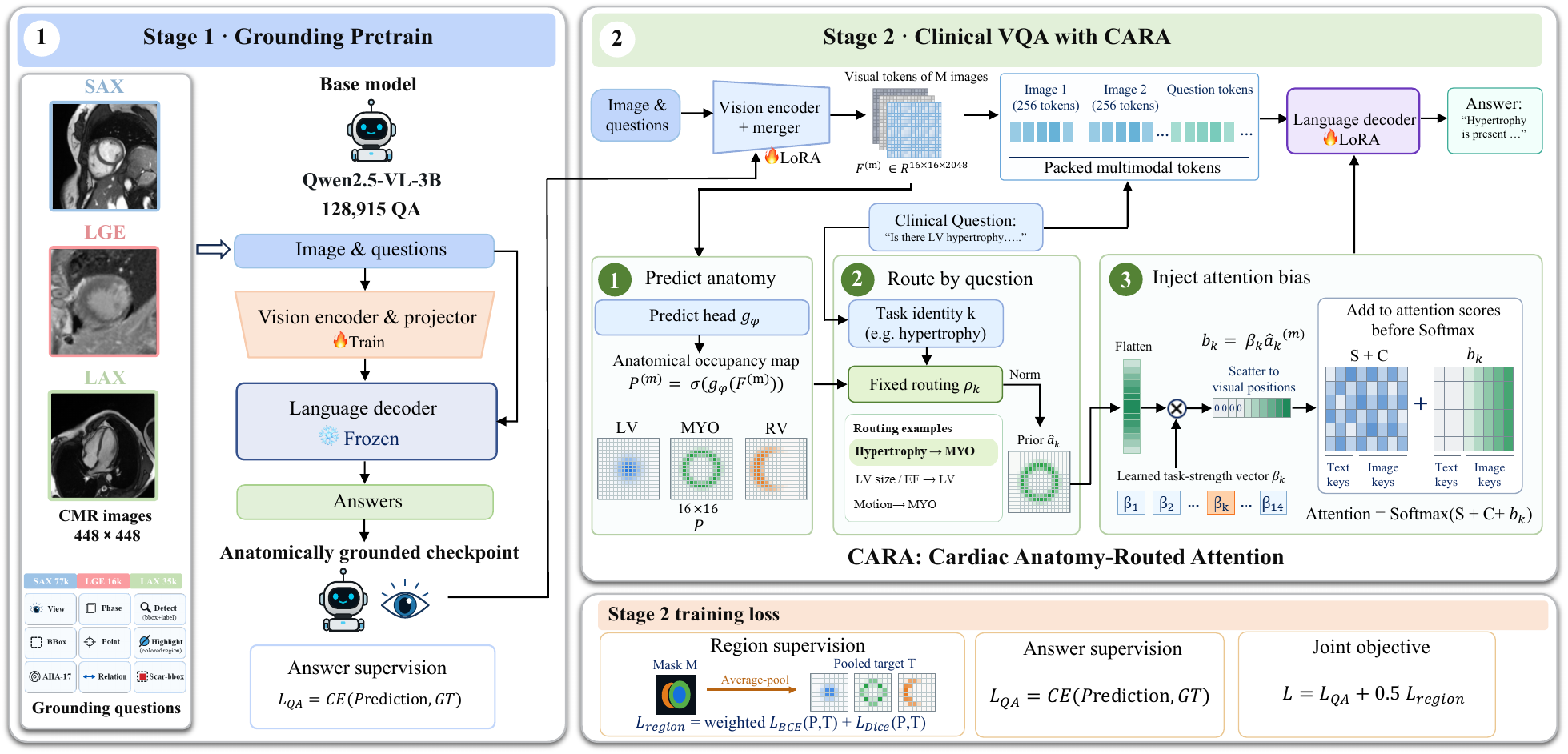}
\caption{Two-stage training with CARA.
Stage~1 learns anatomical grounding while keeping the
language model frozen.
Stage~2 jointly trains clinical answer generation and
anatomical occupancy prediction.
CARA selects the question-relevant occupancy map and
injects the resulting anatomical prior into decoder
attention with a learned task-specific strength.}
\label{fig:two-stage-cara}
\end{figure*}

We present Cardiac Anatomy-Routed Attention (CARA),
which guides decoder attention towards question-relevant
cardiac structures (Figure~\ref{fig:two-stage-cara}).
Below, we detail its anatomical occupancy prediction,
question-guided region routing, and attention injection.



\subsection{Predicting anatomical occupancy}
\label{sec:lattn}

We build CARA on Qwen2.5-VL-3B-Instruct~\citep{bai2025qwen25vltechnicalreport}.
For a QA instance with $M$ input CMR images,
$\mathcal{I}=\{I^{(m)}\}_{m=1}^{M}$, each
$I^{(m)}\in\mathbb{R}^{448\times448\times3}$ is processed by the shared
vision encoder. Qwen2.5-VL uses $14\times14$ image patches and spatially
merges each $2\times2$ group of neighboring patch tokens, yielding a
$16\times16$ grid of post-merger visual representations for each image:
\begin{equation}
F^{(m)} \in \mathbb{R}^{16\times16\times2048},
\qquad m=1,\ldots,M.
\end{equation} 
which are the same visual features passed to the language model. A convolutional head $g_\phi$, shared across images, views, and sequences,
predicts question-independent soft occupancy maps for the LV cavity,
myocardium, and RV cavity:
\begin{equation}
P^{(m)}
=
\sigma\!\left(
g_\phi\!\left(F^{(m)}\right)
\right)
\in [0,1]^{16\times16\times3},
\label{eq:region}
\end{equation}
where $\sigma$ denotes the element-wise sigmoid function. We denote the
occupancy map for structure
$r\in\{\mathrm{LV},\mathrm{MYO},\mathrm{RV}\}$ by
$P_r^{(m)}\in[0,1]^{16\times16}$. The predictor is supervised with soft occupancy targets, where each target
measures the fraction of the corresponding visual-token cell occupied by
the anatomical structure. Thus, $P_r^{(m)}$ provides an image-specific
coarse anatomical prior that is spatially aligned with the VLM visual-token
grid. Segmentation masks are used only to supervise $g_\phi$ during
training; at inference, all occupancy maps are predicted directly from the
input CMR images. Architectural details are provided in
Appendix~\ref{app:cara-architecture}.

\subsection{Anatomy-routed attention injection}
\label{sec:cond}

Each clinical question is associated with one of 14 predefined clinical
tasks, indexed by $k$. For each input image $m$, a fixed routing function
$\rho_k$ selects the anatomical evidence relevant to the task:
\begin{equation}
a_k^{(m)}
=
\rho_k\!\left(
P_{\mathrm{LV}}^{(m)},
P_{\mathrm{MYO}}^{(m)},
P_{\mathrm{RV}}^{(m)}
\right)
\in[0,1]^{16\times16}.
\label{eq:routing}
\end{equation}
For example, hypertrophy-related tasks are routed to the myocardium,
whereas LV-size and EF tasks are routed to the LV cavity. Because occupancy confidence may vary across images, We peak-normalise each routed map independently
to make its scale comparable across images:
\begin{equation}
\widehat a_k^{(m)}
=
\frac{a_k^{(m)}}
{\max\!\left(\epsilon,\left\|a_k^{(m)}\right\|_{\infty}\right)},
\qquad
\epsilon=10^{-6}.
\label{eq:prior}
\end{equation}

Each task has a learned scalar injection strength $\beta_k$. After
flattening the $16\times16$ prior, let $\widehat a_{k,j}^{(m)}$ denote
the value associated with the $j$-th visual token of image $m$, and let
$\pi_m(j)$ map this token to its position in the language-model sequence.
We construct a sequence-aligned bias $b_k\in\mathbb{R}^{L}$ as
\begin{equation}
b_{k,p}
=
\begin{cases}
\beta_k\,\widehat a_{k,j}^{(m)},
& p=\pi_m(j),\\
0,
& p\notin\mathcal I_{\mathrm{visual}},
\end{cases}
\label{eq:bias}
\end{equation}
where $\mathcal I_{\mathrm{visual}}$ denotes the visual-token positions.
This directly aligns each image-specific anatomical prior with its
corresponding visual tokens, including multi-image inputs. CARA then injects this bias into the pre-softmax decoder attention. Let
$S_{i,p}^{(\ell,h)}$ denote the scaled query--key score from query
position $i$ to key position $p$ in decoder layer $\ell$ and head $h$,
and let $C_{i,p}$ denote the original causal and padding mask:
\begin{equation}
\widetilde S_{i,p}^{(\ell,h)}
=
S_{i,p}^{(\ell,h)}
+
C_{i,p}
+
b_{k,p}.
\label{eq:injection}
\end{equation}
Because $b_{k,p}$ depends only on the key position, the same anatomical
bias is shared across decoder layers, heads, and query positions.
Positive $\beta_k$ promotes attention to the routed anatomy, zero
recovers the original attention pattern, and negative $\beta_k$
suppresses it. During autoregressive generation, the bias remains attached
to the original visual tokens and is zero for newly generated positions,
allowing CARA to remain fully mask-free at inference.

\subsection{Two-stage training with anatomical supervision}
\label{sec:training}

Training proceeds from anatomical grounding to clinical VQA.
In Stage~1, we fully fine-tune the vision encoder and
multimodal projector on \nQAone{} anatomical-grounding
QA pairs while keeping the language model frozen.
This checkpoint initialises Stage~2, where we apply LoRA
to adapt the VLM on \nQAtwo{} clinical QA pairs across
14 tasks, jointly optimising CARA's occupancy head
$g_\phi$ and attention strengths $\{\beta_k\}$.

We supervise answer generation with next-token
cross-entropy over answer tokens.
For anatomical supervision, reference masks are
average-pooled onto the visual-token grid to obtain
soft occupancy targets
$R_r^{(m)}\in[0,1]^{16\times16}$ for image $m$
and structure $r$.
The predicted occupancy maps are supervised using
positive-class-weighted binary cross-entropy and soft Dice:
\begin{equation}
\mathcal{L}_{\mathrm{region}}
=
\mathcal{L}_{\mathrm{WBCE}}
+
\mathcal{L}_{\mathrm{Dice}},
\qquad
\mathcal{L}
=
\mathcal{L}_{\mathrm{QA}}
+
\lambda\mathcal{L}_{\mathrm{region}},
\quad \lambda=0.5.
\label{eq:loss}
\end{equation}
Only valid anatomical targets contribute to the region
loss; images without valid masks contribute only to
the QA loss.

Visual features are detached before entering $g_\phi$,
so the region loss does not update the vision encoder.
The QA loss also updates $g_\phi$ and $\beta_k$ through
the differentiable attention bias.
At inference, CARA predicts anatomical priors directly
from visual features without requiring masks as model inputs.
Detailed loss definitions and annotation-validity rules
are provided in Appendix~\ref{app:training}.




\section{Experiments}
\label{sec:exp}

\subsection{Implementation details}
\label{sec:implementation}

\paragraph{Implementation and evaluation.}
\label{sec:evaluation}
We initialise Qwen2.5-VL-3B-Instruct and train the CARA-VL model on four
NVIDIA B200 GPUs using AdamW, bfloat16 precision, and cosine
decay with 5\% warm-up. The base learning rate is $10^{-4}$,
with $10^{-2}$ for Stage~2 CARA injection strengths.
Each stage runs for three epochs, with effective batch
sizes of 64 and 16 for Stage~1 and Stage~2, respectively.
Stage~2 uses LoRA with rank 16, scaling factor 32, and
dropout 0.05. All evaluations use the final Stage~2
checkpoint. Additional training settings are provided
in Appendix~\ref{app:training}.

We evaluate all 14 clinical tasks using the ACDC test
dataset for SAX cine, the M\&Ms-2 test dataset for LAX cine,
and a held-out 50-patient LGE set.
The eight SAX tasks are additionally evaluated on
external In-house B.
Patients are held out across tasks, views, and both
training stages; performance is scored per QA instance. For In-house B, we evaluate the same predictions separately
against two reference standards.
Rule-based labels are generated using the same
measurement-based labelling framework as the training data.
Report-based labels are extracted directly from clinical
reports, with echocardiographic grading thresholds used
for selected findings. Cohort-specific criteria and differences between the two
standards are detailed in Appendix~\ref{app:dd-splits}. For classification, we report balanced accuracy
(BA; mean sensitivity and specificity) and macro-$F_1$.
Regional hypertrophy and wall motion use micro-$F_1$
over question--segment pairs, with each anatomical
segment labelled affected or unaffected.
Scoring details are provided in
Appendix~\ref{app:evaluation-metrics}.

\paragraph{Baselines}
\label{sec:baselines}

We compare our CARA-VL model with general medical VLMs and cardiac-specific
systems. The general medical baselines comprise four open models:
Lingshu-7B~\citep{xu2025lingshu},
HuatuoGPT-Vision-7B~\citep{chen2024huatuogptvision} (Huatuo-Vision),
MedGemma-4B-it~\citep{sellergren2026medgemma}, and
LLaVA-Med-7B~\citep{li2023llava}. All four are evaluated zero-shot,
without adaptation to our CMR training data. The cardiac-specific baselines include
CMR-CLIP~\citep{nakashima2026contrastive}, the contrastive CMR model
of \citet{shad2026generalizable}, and
BAAI Cardiac Agent~\citep{qu2026baai}. For BAAI Cardiac Agent,
we evaluate its VLM component (BAAI-VLM) for direct image-based question
answering without external tool calls.

\suppressfloats[t]
\begin{table}[t]
\centering
\caption{QA-level clinical VQA results on the ACDC test set
(SAX), the 50-patient LGE test set, and the M\&Ms-2 test set
(LAX). Classification cells show binary BA / macro-$F_1$;
LV size, EF, and EDV are evaluated as normal versus abnormal.
Regional tasks ($\dagger$) report micro-$F_1$. All scores are percentages(\%).}
\label{tab:main}
\label{tab:medical_vlm_benchmark}
\begingroup
\CMRTableSetup
\fontsize{8.2}{10.2}\selectfont
\renewcommand{\arraystretch}{1.15}
\newcommand{\CMRPair}[2]{%
  \makebox[1.8em][r]{#1}\hspace{0.18em}/\hspace{0.18em}%
  \makebox[1.8em][r]{#2}}
\newcommand{\CMRSingle}[1]{\makebox[2.8em][r]{#1}}

\begin{tabular*}{\linewidth}{@{\extracolsep{\fill}}l*{7}{c}@{\hspace{6pt}}*{2}{c}@{}}
\toprule[0.85pt]
\multicolumn{10}{@{}l}{\CMRGroup{(a) SAX cine}} \\
\arrayrulecolor{CMRaccent}\cmidrule(lr){1-10}\arrayrulecolor{CMRink}
\textbf{Model}
& \CMRMetric{Hyper.} & \CMRMetric{LV size} & \CMRMetric{RV size}
& \CMRMetric{EF} & \CMRMetric{GRS} & \CMRMetric{GCS}
& \CMRGroup{Avg.}
& \CMRMetric{\shortstack{Reg.\\hyper.$\dagger$}}
& \CMRMetric{\shortstack{Reg.\\motion$\dagger$}} \\
\midrule[0.5pt]
\multicolumn{10}{@{}l}{\CMRGroup{General medical VLMs}} \\
LLaVA-Med-7B & \CMRPair{50.0}{9.9} & \CMRPair{50.0}{24.3} & \CMRPair{50.0}{42.4} & \CMRPair{47.5}{39.9} & \CMRPair{50.0}{41.0} & \CMRPair{50.0}{38.8} & \CMRPair{49.6}{32.7} & \CMRSingle{9.2} & \CMRSingle{31.2} \\
MedGemma-4B-it & \CMRPair{56.6}{55.7} & \CMRPair{60.7}{46.0} & \CMRPair{36.7}{35.5} & \CMRPair{50.0}{27.8} & \CMRPair{50.0}{23.4} & \CMRPair{51.1}{39.4} & \CMRPair{50.9}{38.0} & \CMRSingle{9.2} & \CMRSingle{31.9} \\
Huatuo-Vision & \CMRPair{50.0}{47.1} & \CMRPair{50.0}{24.3} & \CMRPair{50.0}{20.9} & \CMRPair{50.0}{27.8} & \CMRPair{52.1}{35.5} & \CMRPair{59.5}{59.5} & \CMRPair{51.9}{35.9} & \CMRSingle{0.9} & \CMRSingle{24.1} \\
Lingshu-7B & \CMRPair{50.0}{47.1} & \CMRPair{44.4}{27.6} & \CMRPair{50.8}{22.7} & \CMRPair{50.0}{27.8} & \CMRPair{39.2}{39.5} & \CMRPair{45.7}{43.5} & \CMRPair{46.7}{34.7} & \CMRSingle{3.2} & \CMRSingle{16.9} \\
\arrayrulecolor{CMRline}\specialrule{0.4pt}{4pt}{2pt}\arrayrulecolor{CMRink}
\multicolumn{10}{@{}l}{\CMRGroup{CMR-specific models}} \\
CMR-CLIP & \CMRPair{80.0}{60.6} & \CMRPair{62.5}{47.9} & \CMRPair{39.4}{39.7} & \CMRPair{81.4}{77.2} & \CMRPair{65.4}{57.4} & \CMRPair{80.1}{75.6} & \CMRPair{68.1}{59.7} & \CMRSingle{20.3} & \CMRSingle{52.0} \\
\citet{shad2026generalizable} & \CMRPair{42.9}{43.6} & \CMRPair{50.0}{41.9} & \CMRPair{51.1}{44.7} & \CMRPair{50.0}{27.8} & \CMRPair{47.7}{39.8} & \CMRPair{41.1}{37.1} & \CMRPair{47.1}{39.2} & \CMRSingle{14.6} & \CMRSingle{21.9} \\
BAAI-VLM & \CMRPair{49.1}{46.7} & \CMRPair{50.0}{40.4} & \CMRPair{50.0}{42.4} & \CMRPair{64.4}{63.2} & \CMRPair{59.3}{58.5} & \CMRPair{76.7}{77.8} & \CMRPair{58.3}{54.8} & \CMRSingle{11.3} & \CMRSingle{26.6} \\
\arrayrulecolor{CMRline}\cmidrule(lr){1-10}\arrayrulecolor{CMRink}
\CMRGroup{CARA-VL (Ours)} & \CMRPair{\textbf{86.7}}{\textbf{84.3}} & \CMRPair{\textbf{89.9}}{\textbf{88.4}} & \CMRPair{\textbf{86.6}}{\textbf{88.8}} & \CMRPair{\textbf{92.1}}{\textbf{90.4}} & \CMRPair{\textbf{92.6}}{\textbf{92.6}} & \CMRPair{\textbf{96.4}}{\textbf{95.7}} & \CMRPair{\textbf{90.7}}{\textbf{90.0}} & \CMRSingle{\textbf{61.8}} & \CMRSingle{\textbf{55.8}} \\
\bottomrule[0.85pt]
\end{tabular*}

\vspace{2pt}
\begin{tabular*}{\linewidth}{@{\extracolsep{\fill}}l*{6}{c}@{}}
\toprule[0.85pt]
\multirow{2}{*}{\textbf{Model}}
& \multicolumn{3}{c}{\CMRGroup{(b) Late gadolinium enhancement}}
& \multicolumn{3}{c}{\CMRGroup{(c) Long-axis cine}} \\
\arrayrulecolor{CMRaccent}\cmidrule(lr){2-4}\cmidrule(lr){5-7}\arrayrulecolor{CMRink}
& \CMRMetric{Scar presence} & \CMRMetric{Transmurality}
& \CMRMetric{Extent} & \CMRMetric{EF}
& \CMRMetric{EDV} & \CMRMetric{GLS} \\
\midrule[0.5pt]
\multicolumn{7}{@{}l}{\CMRGroup{General medical VLMs}} \\
LLaVA-Med-7B & \CMRPair{50.0}{21.0} & \CMRPair{50.0}{12.2} & \CMRPair{50.0}{40.6} & \CMRPair{50.0}{36.8} & \CMRPair{50.0}{22.0} & \CMRPair{50.0}{41.8} \\
MedGemma-4B-it & \CMRPair{56.7}{57.0} & \CMRPair{53.9}{20.2} & \CMRPair{50.0}{40.6} & \CMRPair{48.9}{36.3} & \CMRPair{48.3}{30.9} & \CMRPair{49.0}{47.1} \\
Huatuo-Vision & \CMRPair{57.5}{48.3} & \CMRPair{50.0}{12.2} & \CMRPair{48.3}{26.0} & \CMRPair{50.0}{29.5} & \CMRPair{50.0}{22.0} & \CMRPair{47.0}{38.0} \\
Lingshu-7B & \CMRPair{50.0}{21.0} & \CMRPair{50.0}{12.2} & \CMRPair{50.0}{40.6} & \CMRPair{50.0}{29.5} & \CMRPair{50.0}{22.0} & \CMRPair{52.8}{51.4} \\
\arrayrulecolor{CMRline}\specialrule{0.4pt}{4pt}{2pt}\arrayrulecolor{CMRink}
\multicolumn{7}{@{}l}{\CMRGroup{CMR-specific models}} \\
CMR-CLIP & \CMRPair{63.1}{63.1} & \CMRPair{56.7}{53.8} & \CMRPair{51.1}{26.5} & \CMRPair{70.0}{64.8} & \CMRPair{54.8}{31.9} & \CMRPair{68.5}{67.6} \\
\citet{shad2026generalizable} & \CMRPair{50.3}{44.3} & \CMRPair{50.0}{12.2} & \CMRPair{52.4}{31.8} & \CMRPair{53.0}{41.9} & \CMRPair{48.2}{46.2} & \CMRPair{30.5}{31.4} \\
BAAI-VLM & \CMRPair{50.1}{49.6} & \CMRPair{43.0}{39.5} & \CMRPair{33.0}{27.4} & \CMRPair{69.3}{68.9} & \CMRPair{57.5}{58.1} & \CMRPair{55.9}{53.2} \\
\arrayrulecolor{CMRline}\cmidrule(lr){1-7}\arrayrulecolor{CMRink}
\CMRGroup{CARA-VL (Ours)} & \CMRPair{\textbf{82.9}}{\textbf{82.0}} & \CMRPair{\textbf{68.5}}{\textbf{71.1}} & \CMRPair{\textbf{82.5}}{\textbf{82.3}} & \CMRPair{\textbf{78.1}}{\textbf{74.9}} & \CMRPair{\textbf{84.9}}{\textbf{81.0}} & \CMRPair{\textbf{77.1}}{\textbf{79.6}} \\

\bottomrule[0.85pt]
\end{tabular*}
\arrayrulecolor{black}
\endgroup
\end{table}

\begin{table}[t]
\centering
\caption{External VQA results on In-house B.
Classification cells show BA / macro-$F_1$; regional cells
show micro-$F_1$ (\%). Avg.\ covers six
classification tasks. Report-based GRS/GCS use
wall-motion-score proxies; reference definitions, regional coverage,
and report-field availability are detailed in Appendix~\ref{app:dd-splits}.}
\label{tab:wu-sax-qa}
\begingroup
\CMRTableSetup
\fontsize{8.2}{10.2}\selectfont
\newcommand{\CMRWUPair}[2]{\makebox[4.5em][c]{#1\,/\,#2}}
\setlength{\tabcolsep}{2pt}
\resizebox{\linewidth}{!}{%
\begin{tabular}{@{}ll*{7}{c}@{\hspace{6pt}}*{2}{c}@{}}
\toprule[0.85pt]
\textbf{Model} & \textbf{Reference}
& \CMRMetric{Hyper.} & \CMRMetric{LV size} & \CMRMetric{RV size}
& \CMRMetric{EF} & \CMRMetric{GRS} & \CMRMetric{GCS}
& \CMRGroup{Avg.}
& \CMRMetric{\shortstack{Reg.\\hyper.}}
& \CMRMetric{\shortstack{Reg.\\motion}} \\
\midrule[0.5pt]
\multicolumn{11}{@{}l}{\CMRGroup{General medical VLMs}} \\
\multirow{2}{*}{LLaVA-Med-7B} & Rule-based & \CMRWUPair{50.0}{13.3} & \CMRWUPair{50.6}{16.2} & \CMRWUPair{50.0}{49.4} & \CMRWUPair{49.0}{47.3} & \CMRWUPair{50.0}{47.6} & \CMRWUPair{50.0}{48.5} & \CMRWUPair{49.9}{37.0} & 7.2 & 10.5 \\
& Report-based & \CMRWUPair{50.0}{23.0} & \CMRWUPair{50.4}{41.2} & \CMRWUPair{50.0}{40.4} & \CMRWUPair{49.6}{49.4} & \CMRWUPair{50.0}{42.8} & \CMRWUPair{50.0}{43.0} & \CMRWUPair{50.0}{40.0} & 32.2 & 28.9 \\
\addlinespace[2pt]
\multirow{2}{*}{MedGemma-4B-it} & Rule-based & \CMRWUPair{58.3}{48.6} & \CMRWUPair{56.1}{43.0} & \CMRWUPair{52.2}{48.9} & \CMRWUPair{49.8}{27.6} & \CMRWUPair{48.9}{13.3} & \CMRWUPair{46.5}{36.1} & \CMRWUPair{51.9}{36.2} & 7.2 & 12.7 \\
& Report-based & \CMRWUPair{52.9}{50.7} & \CMRWUPair{55.9}{55.7} & \CMRWUPair{51.3}{48.8} & \CMRWUPair{50.1}{21.8} & \CMRWUPair{48.0}{23.5} & \CMRWUPair{47.4}{44.0} & \CMRWUPair{50.9}{40.7} & 32.2 & 39.0 \\
\addlinespace[2pt]
\multirow{2}{*}{Huatuo-Vision} & Rule-based & \CMRWUPair{49.5}{48.9} & \CMRWUPair{50.0}{11.1} & \CMRWUPair{50.0}{2.3} & \CMRWUPair{50.0}{24.1} & \CMRWUPair{47.3}{19.3} & \CMRWUPair{55.8}{46.7} & \CMRWUPair{50.4}{25.4} & 3.3 & 9.2 \\
& Report-based & \CMRWUPair{49.5}{45.8} & \CMRWUPair{50.0}{36.9} & \CMRWUPair{50.0}{24.4} & \CMRWUPair{50.0}{17.9} & \CMRWUPair{48.3}{29.7} & \CMRWUPair{49.5}{49.2} & \CMRWUPair{49.6}{34.0} & 5.3 & 26.1 \\
\addlinespace[2pt]
\multirow{2}{*}{Lingshu-7B} & Rule-based & \CMRWUPair{50.0}{45.8} & \CMRWUPair{53.5}{51.3} & \CMRWUPair{50.8}{4.0} & \CMRWUPair{50.0}{24.1} & \CMRWUPair{49.2}{42.5} & \CMRWUPair{49.5}{49.3} & \CMRWUPair{50.5}{36.2} & 1.9 & 8.8 \\
& Report-based & \CMRWUPair{50.0}{41.2} & \CMRWUPair{51.5}{45.6} & \CMRWUPair{49.8}{25.7} & \CMRWUPair{50.0}{17.9} & \CMRWUPair{52.1}{50.1} & \CMRWUPair{50.2}{46.0} & \CMRWUPair{50.6}{37.8} & 22.1 & 25.6 \\
\arrayrulecolor{CMRline}\specialrule{0.4pt}{4pt}{2pt}\arrayrulecolor{CMRink}
\multicolumn{11}{@{}l}{\CMRGroup{CMR-specific models}} \\
\multirow{2}{*}{CMR-CLIP} & Rule-based & \CMRWUPair{85.8}{74.2} & \CMRWUPair{62.5}{34.6} & \CMRWUPair{51.1}{46.6} & \CMRWUPair{\textbf{76.8}}{\textbf{71.7}} & \CMRWUPair{59.8}{28.2} & \CMRWUPair{68.7}{39.2} & \CMRWUPair{67.5}{49.1} & 18.9 & 21.2 \\
& Report-based & \CMRWUPair{\textbf{78.8}}{\textbf{78.3}} & \CMRWUPair{63.4}{\textbf{62.9}} & \CMRWUPair{50.9}{\textbf{49.9}} & \CMRWUPair{\textbf{75.9}}{\textbf{64.0}} & \CMRWUPair{60.9}{43.3} & \CMRWUPair{69.4}{59.1} & \CMRWUPair{\textbf{66.5}}{59.6} & \textbf{57.3} & \textbf{54.4} \\
\addlinespace[2pt]
\multirow{2}{*}{\citet{shad2026generalizable}} & Rule-based & \CMRWUPair{52.9}{41.7} & \CMRWUPair{47.1}{47.3} & \CMRWUPair{50.0}{49.4} & \CMRWUPair{49.9}{24.0} & \CMRWUPair{49.5}{49.4} & \CMRWUPair{39.5}{18.5} & \CMRWUPair{48.2}{38.4} & 10.8 & 8.7 \\
& Report-based & \CMRWUPair{47.3}{43.6} & \CMRWUPair{45.1}{36.0} & \CMRWUPair{50.0}{40.4} & \CMRWUPair{49.9}{17.8} & \CMRWUPair{49.2}{46.3} & \CMRWUPair{46.0}{30.6} & \CMRWUPair{47.9}{35.8} & 29.3 & 18.2 \\
\addlinespace[2pt]
\multirow{2}{*}{BAAI-VLM} & Rule-based & \CMRWUPair{50.0}{46.1} & \CMRWUPair{50.0}{46.7} & \CMRWUPair{50.0}{49.4} & \CMRWUPair{51.9}{45.1} & \CMRWUPair{54.3}{55.4} & \CMRWUPair{58.2}{53.0} & \CMRWUPair{52.4}{49.3} & 5.9 & 12.5 \\
& Report-based & \CMRWUPair{49.9}{41.4} & \CMRWUPair{50.0}{29.4} & \CMRWUPair{50.0}{40.4} & \CMRWUPair{52.7}{49.7} & \CMRWUPair{52.7}{49.5} & \CMRWUPair{57.5}{58.1} & \CMRWUPair{52.1}{44.7} & 28.9 & 34.4 \\
\addlinespace[2pt]
\arrayrulecolor{CMRline}\cmidrule(lr){1-11}\arrayrulecolor{CMRink}
\multirow{2}{*}{\CMRGroup{CARA-VL (Ours)}} & Rule-based & \CMRWUPair{\textbf{87.1}}{\textbf{90.7}} & \CMRWUPair{\textbf{82.5}}{\textbf{66.6}} & \CMRWUPair{\textbf{62.5}}{\textbf{59.1}} & \CMRWUPair{73.5}{66.3} & \CMRWUPair{\textbf{87.3}}{\textbf{86.0}} & \CMRWUPair{\textbf{84.1}}{\textbf{65.8}} & \CMRWUPair{\textbf{79.5}}{\textbf{72.4}} & \textbf{55.1} & \textbf{36.7} \\
& Report-based & \CMRWUPair{69.9}{72.7} & \CMRWUPair{\textbf{64.0}}{60.7} & \CMRWUPair{\textbf{52.4}}{46.8} & \CMRWUPair{72.6}{58.2} & \CMRWUPair{\textbf{62.7}}{\textbf{64.5}} & \CMRWUPair{\textbf{75.0}}{\textbf{77.2}} & \CMRWUPair{66.1}{\textbf{63.4}} & 30.9 & 34.5 \\
\bottomrule[0.85pt]
\end{tabular}}
\arrayrulecolor{black}
\endgroup
\end{table}

\subsection{Main results}
\label{sec:medical-vlm-comparison}

\paragraph{Limited zero-shot generalization of general medical VLMs.}
General medical VLMs struggle with fine-grained CMR
clinical QA (Tables~\ref{tab:main} and~\ref{tab:wu-sax-qa}).
Their mean SAX BA ranges from 46.7\% to 51.9\%,
close to the binary chance level.
Performance on LGE and LAX is also generally limited.
These results indicate that broad medical pretraining
alone does not reliably support the cardiac visual
recognition and localisation required by our tasks.

\paragraph{Internal evaluation: clinical assessment and localisation.}
CARA-VL achieves the highest scores across all 14 internal
tasks, leading in both BA and macro-$F_1$ for classification
and micro-$F_1$ for regional localisation
(Table~\ref{tab:main}).
On SAX, it reaches 90.7\% mean BA and 90.0\% mean macro-$F_1$,
exceeding CMR-CLIP by 22.6 and 30.3 percentage points,
respectively. Its advantage extends to structural,
functional, and scar assessment in LAX and LGE.
Regional hypertrophy shows a particularly pronounced gain:
61.8\% micro-$F_1$ versus 20.3\% for the strongest baseline;
regional wall motion reaches 55.8\% versus 52.0\%.
These gains extend beyond global abnormality classification
to identifying affected myocardial regions, supporting
fine-grained clinical assessment across CMR imaging settings.

\paragraph{External evaluation: generalization across cohorts and clinical standards.}
On In-house B, CARA-VL leads under Rule-based labels
on five of six classification tasks and both regional
tasks (Table~\ref{tab:wu-sax-qa}).
Its mean BA / macro-$F_1$ reaches 79.5\% / 72.4\%,
compared with 67.5\% / 49.1\% for CMR-CLIP,
with EF remaining the exception.
Results are more mixed under Report-based labels:
CMR-CLIP performs better on hypertrophy and EF,
and achieves regional hypertrophy and wall-motion
micro-$F_1$ of 57.3\% and 54.4\%, compared with
30.9\% and 34.5\% for CARA-VL.
CMR-CLIP's report-supervised pretraining may contribute
to its stronger agreement with these labels.
These results support generalization to an unseen cohort,
while highlighting a remaining gap in agreement with
clinical reporting.
The contrast between the two reference standards also
shows that external performance depends on how clinical
findings are defined and recorded.
Differences in reference criteria and anatomical coverage
are detailed in Appendix~\ref{app:dd-splits}
(Table~\ref{tab:wu-label-differences}).

\subsection{Ablation}
\label{sec:cond-results}
\begin{figure}[htbp]
\centering
\includegraphics[width=\linewidth]{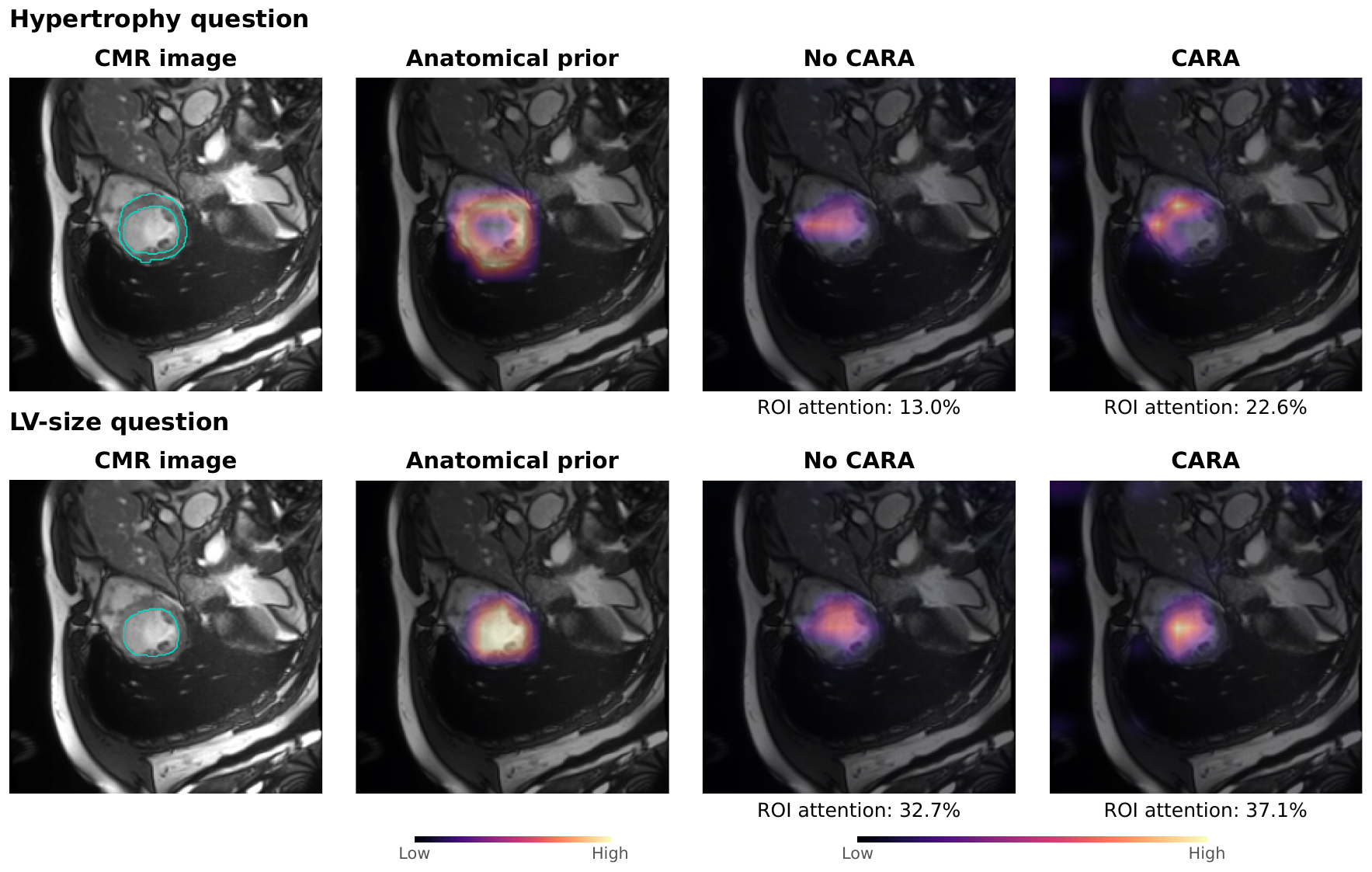}
\caption{Attention maps from models trained without and
with CARA for hypertrophy and LV-size questions on the
same SAX image.
Attention maps average all heads in the final eight
decoder layers at the last prompt token.
Attention maps share a colour scale within each row;
anatomical priors are separately normalised to unit peak.
ROI attention quantifies the fraction of visual attention
allocated to the target anatomy.}
\label{fig:cara-attention-comparison}
\end{figure}

\begingroup
\setlength{\intextsep}{5pt}
\setlength{\columnsep}{11pt}
\begin{wraptable}{r}{0.6\textwidth}
\centering
\captionsetup{font=footnotesize,skip=4pt}
\caption{Ablation on ACDC SAX clinical QA after Stage-2
training. BA and macro-$F_1$ are averaged across the six
global classification tasks (\%).}
\label{tab:lattn-primary}
\CMRTableSetup
\fontsize{8}{9.2}\selectfont
\setlength{\tabcolsep}{1.5pt}
\renewcommand{\arraystretch}{1.10}
\begin{tabular*}{\linewidth}{@{\extracolsep{\fill}}lcccc@{}}
\toprule[0.7pt]
\textbf{Model}
& \CMRMetric{Stage-1}
& \CMRMetric{CARA}
& \CMRMetric{BA}
& \CMRMetric{Macro-$F_1$} \\
\midrule[0.4pt]
\CMRGroup{CARA-VL (Ours)}
& $\checkmark$ & $\checkmark$
& 90.7 & 90.0 \\
\quad w/o Stage-1
& $\times$ & $\checkmark$
& 88.4 & 87.9 \\
\quad w/o CARA
& $\checkmark$ & $\times$
& 86.8 & 86.5 \\
\quad w/o Stage-1 and CARA
& $\times$ & $\times$
& 86.2 & 86.0 \\
\bottomrule[0.7pt]
\end{tabular*}
\arrayrulecolor{black}
\end{wraptable}
Table~\ref{tab:lattn-primary} evaluates the contributions
of anatomical grounding pretraining and CARA.
Removing Stage-1 reduces mean BA and macro-$F_1$ by
2.3 and 2.1 percentage points, respectively, while removing
CARA produces larger drops of 3.9 and 3.5 points.
The full model outperforms the variant without either
component by 4.5 points in BA and 4.0 points in macro-$F_1$.
Stage-1 also provides a larger gain when CARA is enabled,
suggesting that anatomical grounding and guided attention
offer complementary benefits.
Figure~\ref{fig:cara-attention-comparison} compares attention
maps from models trained without and with CARA.
For the same SAX image, the model trained with CARA shows
more concentrated attention on the myocardium for
hypertrophy and the LV cavity for LV size.
The alignment between the routed priors and decoder
attention provides a qualitative view of CARA's
anatomical guidance.
These examples illustrate how anatomical routing guides
attention towards different cardiac structures according
to the clinical question.
\par
\endgroup

\section{Discussion}
\label{sec:disc}

Our study demonstrates the value of anatomical supervision
for developing and evaluating cardiac VLMs.
Our dataset covers 14 clinical QA tasks across three CMR
imaging modalities, deriving reference answers from segmentation
masks, quantitative measurements, and task-specific criteria.
These tasks assess cardiac structure, function, tissue
characteristics, and regional localisation.
CARA connects anatomical grounding with anatomy-routed
attention, using learned anatomical priors to guide image interpretation.
Its leading performance across all 14 internal tasks,
together with its performance on an external clinical cohort,
demonstrates capabilities in global abnormality recognition
and regional assessment.
Together, the data and method provide a practical
framework for studying cardiac visual understanding
and highlight anatomical learning as a promising
direction for specialised medical VLMs.

\paragraph{Limitations.}
Our study has three main limitations.
First, training answers are generated from annotation masks
and image-derived measurements using predefined rules,
inheriting errors and assumptions from these procedures.
Although useful for evaluating visual recognition and
localisation, these labels require independent clinical
validation, as differences between measurement-based
and report-based evaluation further highlight
(Appendix~\ref{app:dd-splits}).
Second, the underlying cohorts remain modest:
generating multiple questions per examination increases
QA volume without proportionally increasing patient diversity.
Access to CMR studies, usable annotations, and quality review
constrains expansion, while external evaluation currently
covers SAX cine only.
Finally, current performance remains insufficient for
clinical use, and answering predefined questions from
selected images captures only part of comprehensive
CMR interpretation. If larger multicentre CMR datasets with paired clinical
reports become available, future work could incorporate
report supervision and assess clinical utility through
independent expert review.

\section{Conclusion}
We presented CARA-VL, which combines anatomical grounding
and anatomy-routed attention for fine-grained CMR question
answering.
Our data pipeline connects clinical reporting needs with
anatomical annotations and quantitative measurements,
providing grounding and clinical QA supervision across
SAX cine, LGE, and LAX cine without paired reports for individual training images.
CARA-VL leads the evaluated models across all 14 internal
tasks, with ablations supporting the complementary benefits
of anatomical learning and guided attention.
External evaluation shows generalization to an unseen clinical
cohort alongside remaining gaps in agreement with clinical reporting.
Future work could extend anatomical guidance to full cine
sequences, linking cardiac structure with motion over time,
and explore more open-ended clinical questions.
These directions provide opportunities to advance generative
cardiac VQA and study how anatomical knowledge supports
visual understanding in medical VLMs.

\subsubsection*{Reproducibility statement}
We provide detailed descriptions of data construction,
model implementation, training, and evaluation in
Appendices~\ref{app:datadetails} and~\ref{app:training}
to facilitate reproducibility.
We will release a subset of the QA data constructed
from public CMR datasets, in accordance with their
licences and redistribution conditions, to support
further research on cardiac visual question answering.


\subsubsection*{AI use statement}
Generative AI tools assisted with manuscript drafting,
language editing and translation, literature search,
code development and review, figure preparation, and
interpretation and consistency checking of reported results.
The authors take responsibility for the final manuscript
and associated research artifacts, including the scientific
claims, references, implementation, and reported results.

\bibliography{iclr2027_conference}
\bibliographystyle{iclr2027_conference}
\clearpage
\appendix

\section{Related work}
\label{sec:related}

\paragraph{General medical VQA models.}
General medical VLMs such as LLaVA-Med~\citep{li2023llava}, Med-Flamingo~\citep{moor2023med}, HuatuoGPT-Vision~\citep{chen2024huatuogptvision}, and MedGemma~\citep{sellergren2026medgemma} have demonstrated strong medical visual question answering capabilities through large-scale multimodal pretraining and medical instruction tuning.
However, their visual training is largely built around broad biomedical figures and common medical imaging domains, including radiography, pathology, dermatology, and ophthalmology~\citep{lu2025integrating, li2025vision}, rather than systematic supervision on cardiac MRI.
This specialization gap is particularly consequential for CMR, where clinically meaningful answers depend on cardiac anatomy, cine dynamics, physical-scale measurements, and sequence-specific pathology~\citep{schulz2020standardized, rajiah2023cardiac}.
Thus, strong general medical VQA capability does not necessarily translate to reliable CMR-specific clinical assessment.

\paragraph{Emerging CMR-specific foundation and vision--language models.}
Recent work has increasingly explored foundation and multimodal models tailored to cardiac MRI.
CMR-CLIP~\citep{nakashima2026contrastive} and ~\cite{shad2026generalizable} both adopt CLIP-style image--text representation learning from cine CMR and associated clinical reports, targeting zero-shot disease recognition and downstream functional or diagnostic prediction. CineMA~\citep{fu2026development} scales self-supervised cine-CMR pretraining to support tasks including segmentation, landmark localization, diagnosis, and prognosis, while MARCUS~\citep{o2026marcus} introduces modality-specific cardiac vision--language experts for ECG, echocardiography, and CMR within an agentic multimodal system. BAAI Cardiac Agent~\citep{qu2026baai} combines CMR-specific
VQA with expert-model orchestration for cardiac segmentation,
functional quantification, disease diagnosis, and structured
report generation. Together, these studies highlight growing interest in CMR-specific VLMs, but current approaches often emphasize contrastive representation learning, rely on heterogeneous or complex input settings, and lack a standardized fine-grained clinical QA formulation. Their dependence on large institutional or non-public cohorts also limits reproducibility and direct comparison.

\section{Clinical Background on Cardiac MRI}
\label{app:clinical-background}

This appendix introduces cardiac anatomy, common cardiac MRI
views and sequences, and the clinical concepts used to describe
cardiac structure, function, and tissue characteristics. The left and right
ventricular (LV and RV) cavities contain blood, while the
myocardium forms the muscular walls responsible for
contraction.

\subsection{Imaging views and sequences}

\begin{figure*}[t]
    \centering
    \includegraphics[width=0.98\textwidth]{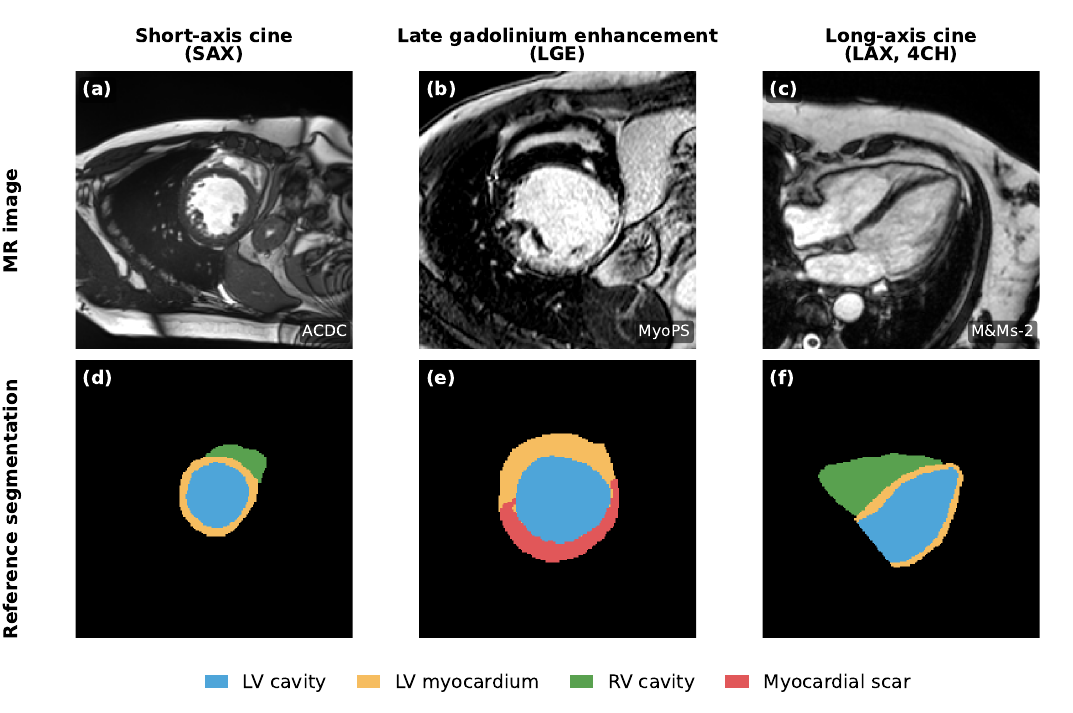}
    \caption{Representative CMR views and spatially matched reference
    annotations. Top: end-diastolic SAX cine (ACDC), short-axis LGE with
    myocardial scar (MyoPS), and end-diastolic four-chamber LAX cine (M\&Ms-2).
    Bottom: the corresponding segmentations. SAX shows ventricular
    cross-sections, LAX displays chamber geometry along the ventricular long
    axis, and LGE delineates hyperenhanced scar. Colours denote the LV cavity
    (blue), LV myocardium (orange), RV cavity (green), and myocardial scar
    (red). These are reference annotations, not model predictions.}
    \label{fig:cmr-views-anatomy}
\end{figure*}

Figure~\ref{fig:cmr-views-anatomy} contrasts the imaging geometry and
reference anatomy used across the three task families.

\paragraph{Short-axis cine.}
Short-axis (SAX) cine imaging provides cross-sectional
views of the ventricles perpendicular to the left
ventricular long axis. Cine acquisitions resolve changes
in ventricular cavity size and myocardial wall thickening
throughout the cardiac cycle. A basal-to-apical stack
enables quantification of ventricular volumes, ejection
fraction, and myocardial mass, while regional assessment
characterises wall thickness and systolic wall motion
\citep{kramer2020standardized,schulz2020standardized}.

\paragraph{Long-axis cine.}
Long-axis (LAX) cine imaging comprises standard two-,
three-, and four-chamber views aligned with the left
ventricular long axis. These planes delineate the
ventricular apex, atrioventricular junctions, and chamber
geometry, providing complementary assessment of global
and regional systolic function. Temporal changes in
myocardial position and ventricular length also support
assessment of longitudinal shortening and mitral annular
excursion
\citep{kramer2020standardized,schulz2020standardized}.

\paragraph{Late gadolinium enhancement.}
Late gadolinium enhancement (LGE) imaging characterises
myocardial injury using delayed contrast-enhanced
acquisitions, with normal myocardial signal suppressed
to delineate hyperenhanced tissue. The location,
distribution, and transmural extent of enhancement
inform the assessment of myocardial scar and the
differentiation of ischaemic and non-ischaemic injury
patterns \citep{schulz2020standardized}.

\subsection{Cardiac structure, function, and tissue characteristics}

The reference masks in Figure~\ref{fig:cmr-views-anatomy} illustrate the
ventricular cavities, surrounding LV myocardium, and myocardial scar.

\paragraph{Wall thickness and chamber size.}
Myocardial hypertrophy is characterised by increased
wall thickness, which may be diffuse or confined to
specific myocardial regions. Ventricular dilation
instead refers to enlargement of the ventricular cavity.
Wall thickness and cavity size therefore describe
distinct aspects of cardiac morphology: a thickened
myocardial wall does not necessarily imply an enlarged
chamber. Their assessment requires clearly defined
imaging planes and cardiac phases
\citep{schulz2020standardized}.

\paragraph{AHA 17-segment model.}
The American Heart Association (AHA) 17-segment model
standardises the anatomical subdivision of the left
ventricular myocardium \citep{aha17}.
It comprises six basal segments, six mid-ventricular
segments, four apical segments, and the apical cap.
The basal and mid-ventricular levels each include
anterior, anteroseptal, inferoseptal, inferior,
inferolateral, and anterolateral segments, whereas
the apical level contains anterior, septal, inferior,
and lateral segments. The apical cap represents
the terminal myocardium beyond the ventricular cavity.
This nomenclature specifies both the base-to-apex
level and circumferential position, providing a
consistent reference for reporting regional
myocardial findings.

\paragraph{Ventricular volume and systolic function.}
In cine CMR, end-diastole and end-systole are conventionally
identified as the phases of maximal and minimal ventricular
cavity volume, respectively. The corresponding volumes
are termed end-diastolic volume (EDV) and end-systolic
volume (ESV). Their difference defines stroke volume,
and ejection fraction (EF) expresses stroke volume
as a percentage of EDV
\citep{schulz2020standardized}:
\[
\mathrm{EF}
=
\frac{\mathrm{EDV}-\mathrm{ESV}}{\mathrm{EDV}}
\times 100\%.
\]
EDV characterises ventricular cavity size and is commonly
indexed to body surface area for comparison with normal
reference ranges \citep{kawel2020reference}.
EF provides a global measure of systolic emptying,
whereas regional wall-motion assessment evaluates
myocardial displacement and systolic thickening
within individual segments. Beyond these end-diastolic
and end-systolic measures, ventricular volume--time
curves further characterise the timing and rates of
ventricular emptying and filling \citep{axel2026analysis}.

\paragraph{Myocardial deformation.}
Myocardial strain quantifies tissue deformation relative
to a reference configuration, conventionally end-diastole,
and characterises regional and global myocardial mechanics
\citep{pedrizzetti2016principles}.
Global radial strain (GRS) describes myocardial wall
thickening, whereas global circumferential strain (GCS)
and global longitudinal strain (GLS) describe shortening
around the ventricular circumference and along the
base-to-apex axis, respectively. ``Global'' refers to
an aggregate measure over the analysed myocardium.
Under the conventional sign convention, systolic radial
strain is positive, while circumferential and longitudinal
strain are negative; impaired shortening is therefore
reflected by less negative values.
CMR-based deformation assessment includes feature tracking
of routine cine images and motion estimation from tagged
MRI. Strain estimates depend
on the segmentation procedure, slice coverage, and analysis
software, making consistent measurement definitions
important for interpretation \citep{lim2021quantification, ye2024learning}.

\paragraph{Scar presence, transmurality, and extent.}
Myocardial scar represents replacement of injured
myocardium by fibrotic tissue and can be assessed
using LGE imaging. Scar presence indicates whether
scar is identified within the myocardium.
Transmurality describes the fraction of local myocardial
wall thickness occupied by scar, ranging from
partial-thickness to full-thickness involvement.
Scar extent describes the amount or spatial spread
of affected myocardium \citep{schulz2020standardized}.
Circumferential extent describes the proportion of the
myocardial circumference affected by scar.
Transmurality and circumferential extent thus describe
different dimensions of scar involvement: a lesion
may span the full wall thickness while remaining
confined to a small circumferential region.
Segmentation of scar and myocardial boundaries supports
quantitative assessment of these properties, as explored
in automated scar analysis \citep{papetti2023accurate, guo2025verse}.

\section{Data construction details}
\label{app:datadetails}

We build two training sets from one rendering pipeline (Fig.~\ref{fig:datapipe}). The \stageone{} grounding set reads its answers directly from expert reference masks; the \stagetwo{} clinical set computes its answers from physical measurements on reference or automatically generated masks. Table~\ref{tab:sources} lists every source, its mask provenance, and its role in each stage.

\subsection{From free-text reports to a closed question set}
\label{app:dd-questions}
The 14 questions are not chosen a priori. They are the surviving intersection of what clinical reports actually state, what has a guideline-anchored threshold, and what can be measured from segmentation.

\paragraph{Report corpus and recurring findings.}
We analyse 3{,}215 de-identified CMR reports from three
sources: 1{,}112 disease-labelled reports, 1{,}902 consecutive
clinical reports, and 201 reports accompanied by structured
post-processing results. A curated bilingual lexicon covering
20 clinical finding categories identifies whether each finding
is mentioned in a report. Frequently discussed findings include
myocardial enhancement, chamber dilation, reduced ejection
fraction, wall thickening, and regional wall-motion abnormalities.
These frequencies characterise recurring clinical concerns
and inform candidate question selection.
However, paired CMR images are unavailable for this report
corpus. The reports therefore guide the scope of clinical
questions rather than provide image--question--answer
training pairs.

\begin{figure}[t]
\centering
\includegraphics[width=0.836\linewidth]{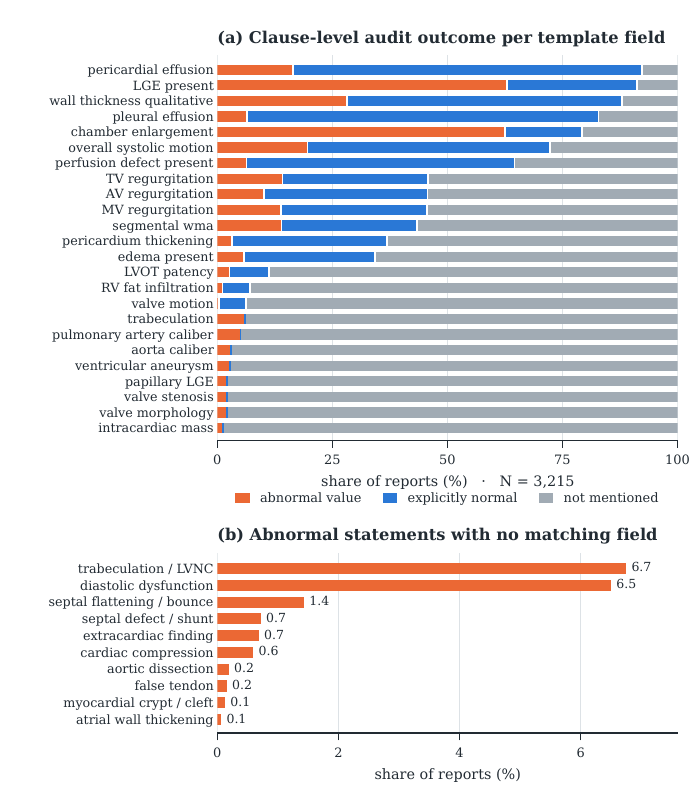}
\caption{Structured report audit over 3,215 CMR reports. \textbf{(a)}~For each template field, the share of reports in which the field is not mentioned, explicitly normal, or carries an abnormal value, ordered by mention rate (explicitly normal $+$ abnormal). A high mention rate with a low abnormality rate marks a finding that radiologists address routinely but rarely find abnormal. \textbf{(b)}~Abnormal statements matching no template field.}
\label{fig:report-audit}
\end{figure}

\paragraph{Structured report audit.}
We organise report findings into a structured template,
with each field assigned an imaging sequence or study-level
scope and a clinical information type, such as a measurement,
visual finding, or anatomical location.
A clause-level audit distinguishes findings that are
not mentioned, explicitly normal, or abnormal, retaining
reported measurements or grades where available.
We iteratively review abnormal statements not captured
by the template and add or refine fields to improve coverage.
The resulting template contains 173 fields, including
59 with predefined answer categories.
We then aggregate mention and abnormality frequencies
separately, distinguishing frequently discussed findings
from frequently reported abnormalities.
Together, these statistics inform candidate question
selection; the template serves as a report-analysis
framework rather than an image-level labelling scheme.
Figure~\ref{fig:report-audit} plots the audit.

\paragraph{Selecting clinical QA tasks.}
Candidate tasks are selected according to three criteria:
they address recurring clinical findings in the reports,
admit explicit measurement or grading criteria, and
are supported by the imaging data and segmentation
annotations available in our cohorts.
Reports guide question selection, while reference answers
are derived separately from image-based measurements
and acquisition metadata. Valvular disease, perfusion defects, oedema, atrial findings,
and parametric mapping are excluded because the available
data do not support the required measurements.
We also exclude ischaemic versus non-ischaemic enhancement
classification because the available LGE cohort lacks
sufficient variation for this distinction, and atrial-plane
images without assessable ventricular myocardium.
The resulting 14 clinical QA tasks are summarised in
Table~\ref{tab:channels}, with grading criteria detailed below.

\paragraph{Task-specific grading criteria.}
We derive categorical reference answers from image-based
measurements using task-specific assessment criteria,
informed by published clinical standards and the definitions
of the measured quantities.
For selected tasks, borderline measurements are excluded
to reduce label ambiguity.
Table~\ref{tab:channels} summarises the measurement rules,
answer categories, and exclusion intervals used for
clinical QA construction.

\begin{table}[t]
\centering
\scriptsize
\setlength{\tabcolsep}{3pt}
\renewcommand{\arraystretch}{1.15}
\caption{Definitions and labelling criteria for the 14 clinical
QA tasks. ED and ES denote end-diastole and end-systole,
respectively. The table summarises answer categories,
measurement rules, and exclusion intervals. Segmentation
sources, quality control, measurement derivation, and
training-set statistics are detailed in
Appendix~\ref{app:dd-stage2}.}
\label{tab:channels}

\begin{tabular}{@{}
p{.07\linewidth}
p{.25\linewidth}
p{.09\linewidth}
p{\dimexpr.59\linewidth-6\tabcolsep\relax}
@{}}
\toprule
Domain & Task and answer categories & Frames
& Measurement and labelling rule \\
\midrule

SAX
& Hypertrophy: present / absent
& ED
& Present if any AHA segment has ED wall thickness
$\ge 13\,\mathrm{mm}$; source-labelled normal cases
are assigned negative.
\\

& Regional hypertrophy: segment list
& ED
& Segments with ED wall thickness $\ge 13\,\mathrm{mm}$;
six segments at basal and mid-ventricular levels,
four at the apical level.
\\

& LV size: normal / mild / moderate / severe dilation
& ED+ES
& Patient-level LV EDV:
$<200$, $[200,240)$, $[240,300)$,
and $\ge 300\,\mathrm{mL}$, respectively.
\\

& EF: normal / mild / moderate / severe reduction
& ED+ES
& Whole-stack area-ratio EF:
$\ge 50\%$, $[40,50)\%$, $[30,40)\%$,
and $<30\%$, respectively.
\\

& RV size: dilated / normal
& ED+ES
& Dilated if RV EDV $\ge 200\,\mathrm{mL}$
and RV/LV EDV ratio $\ge 1.3$;
single-criterion positives require visual confirmation.
Normal if both are below threshold;
unconfirmed candidates are excluded.
\\

& GRS surrogate: normal / reduced
& ED+ES
& Mean systolic wall thickening:
$\ge 45\%$ normal, $<35\%$ reduced;
exclude $[35,45)\%$.
\\

& GCS: normal / reduced
& ED+ES
& Feature-tracking peak circumferential strain:
$\le -16\%$ normal, $\ge -12\%$ reduced;
exclude $(-16,-12)\%$.
\\

& Regional motion: segment list
& ED+ES
& Segments with systolic wall thickening $<20\%$.
Inherits GRS sample filtering, with no additional
segment-level exclusion interval.
\\

\midrule

LGE
& Scar presence: present / absent
& Single
& Present if the reference scar mask is non-empty.
\\

& Transmurality: transmural / non-transmural
& Single
& Transmural if any sliding $60^\circ$ sector has mean
transmural extent $\ge 50\%$.
Automatically labelled positives are retained only
at $\ge 60\%$; recorded review overrides are retained.
\\

& Extent: localised / widespread
& Single
& Widespread if the circumferential scar fraction
is $\ge 0.5$.
Automatically exclude $[0.40,0.60)$;
recorded review overrides are retained.
\\

\midrule

LAX
& EF: normal / mild / moderate / severe reduction
& ED+ES
& Patient-level SAX EF grade where available;
otherwise derived from Kaggle ventricular volumes
or M\&Ms-2 SAX reference masks.
\\

& EDV: normal / mild / moderate / severe dilation
& ED+ES
& Patient-level SAX LV-size grade where available;
otherwise derived from Kaggle EDV
or M\&Ms-2 SAX reference masks.
\\

& GLS: normal / reduced
& ED+ES
& Feature-tracking GLS:
$\le -16\%$ normal, $\ge -12\%$ reduced;
exclude $(-16,-12)\%$.
\\

\bottomrule
\end{tabular}

\end{table}

\subsection{Data Sources}
\label{app:dd-sources}
The corpus spans SAX cine, LGE, and LAX cine (two- and
four-chamber views), with source contributions summarised
in Table~\ref{tab:sources}. Public datasets include
ACDC~\citep{bernard2018deep}, M\&Ms,
M\&Ms-2~\citep{campello2021multi, martin2023deep}, MyoPS-related cohorts~\citep{zhuang2018multivariate,
qiu2023myops,ding2023aligning,ding2025cinemyops},
EMIDEC~\citep{lalande2020emidec}, the cine and LGE subsets
of CMR-MULTI~\citep{qu2026baai}, and the Kaggle Data Science
Bowl Challenge~\citep{kaggle2015cardiac}.
In-house A is an image-only training collection comprising an
expert-annotated cine subset, a clinical image catalogue, and
additional cine collections with automatically generated masks.
In-house B is the image--report cohort used exclusively for external
evaluation. Its final SAX set contains 612 examinations from
461 distinct patient identifiers and provides 14{,}336 SAX QA pairs. We use expert reference segmentation masks where available.
For images without reference masks, we obtain masks from a
deployed internal CMR segmentation pipeline and review their quality
before clinical QA construction.
Table~\ref{tab:sources} summarises the data sources and their
contributions to each training stage and evaluation set.
Segmentation and quality-control procedures are detailed
in Appendix~\ref{app:dd-stage2}, and data-split procedures
in \S\ref{app:dd-splits}.
In-house B is reserved exclusively for external evaluation.

\begin{table}[htbp]
\centering\scriptsize
\setlength{\tabcolsep}{2.5pt}
\caption{Actual source contributions to training and held-out evaluation.}
\label{tab:sources}
\renewcommand{\arraystretch}{1.12}
\begin{tabular}{@{}p{.06\linewidth}p{.24\linewidth}p{.14\linewidth}p{.19\linewidth}p{.19\linewidth}p{\dimexpr.18\linewidth-10\tabcolsep\relax}@{}}
\toprule
Domain & Source & Mask source & \stageone{} & \stagetwo{} train & Held out \\
\midrule
\multicolumn{6}{@{}l}{\textit{Public datasets}} \\
SAX & ACDC & Expert & 100 (8{,}845) & 89 (1{,}803) & 50 \\
 & M\&Ms & Expert & 320 (26{,}846) & 281 (4{,}815) & -- \\
 & M\&Ms-2 & Expert & 200 (16{,}449) & 200 (3{,}841) & 160 \\
 & CMR-MULTI cine & Expert & 105 (10{,}093) & 95 (1{,}459) & -- \\
 & MyoPS-related bSSFP cine & Auto. + QC & -- & 64 (994) & -- \\
 & Kaggle DSB & Auto.\ + QC & -- & 769 (6{,}924) & -- \\
\addlinespace
LGE & MyoPS-related cohorts & Expert & 183 (10{,}900) & 189 (3{,}618) & 30 \\
 & EMIDEC & Expert & 67 (2{,}815) & 86 (1{,}143) & 14 \\
 & CMR-MULTI LGE & Expert & 31 (2{,}678) & 41 (884) & 6 \\
\addlinespace
LAX & M\&Ms-2 (4CH) & Expert & 200 (16{,}800) & 200 (554) & 160 \\
 & CMR-MULTI cine (4CH, 2CH) & Expert & 240 series (18{,}240) & 175 series (175) & -- \\
 & Kaggle DSB & Auto.\ + QC & -- & 1{,}064 (4{,}930) & -- \\
\cmidrule(l){2-6}
 & \multicolumn{2}{@{}l}{Public subtotal (QA)} & 113{,}666 & 31{,}140 & \\
\midrule
\multicolumn{6}{@{}l}{\textit{Internal datasets}} \\
SAX & In-house A & Expert / Auto.\ + QC & 153 (15{,}249) & 9{,}278 QA & -- \\
LAX & In-house A & Auto.\ + QC & -- & 2{,}381 QA & -- \\
SAX & In-house B (with reports) & Auto.\ + QC & -- & -- & 612 exams; 461 patients \\
\cmidrule(l){2-6}
 & \multicolumn{2}{@{}l}{In-house subtotal (QA)} & 15{,}249 & 11{,}659 & \\
\midrule
\multicolumn{3}{@{}l}{Total QA} & \nQAone{} & \nQAtwo{} & \\
\bottomrule
\end{tabular}
\smallskip
\begin{minipage}{\linewidth}
\scriptsize
\textit{Notes.}
Counts are not additive across stages or views.
In-house A clinical training entries report QA counts;
CMR-MULTI LAX entries report series rather than unique patients.
Only expert masks are used for \stageone{}.
\end{minipage}
\end{table}

\subsection{Physical-scale rendering}
\label{app:dd-space}

We standardise spatial resolution and intensity range
before QA generation. Using acquisition-specific pixel
spacing, we resample SAX cine to 1.4\,mm/pixel and LGE
and LAX cine to 1.0\,mm/pixel, with bilinear interpolation
for images and nearest-neighbour interpolation for masks.
Intensities are clipped to the 1st--99th percentiles for
cine and the 2nd--98th percentiles of foreground pixels
for LGE. Images and masks are cropped to $224\times224$
pixels, with segmentation-guided centring where needed
to preserve cardiac coverage. Cases with unresolved
spacing inconsistencies or truncated anatomy are excluded. The cropped images define the reference coordinate system
and are subsequently upsampled bicubically to
$448\times448$ for model input, with spatial coordinates
scaled accordingly. All compared models receive the
same preprocessed images.

\subsection{Anatomical grounding supervision}
\label{app:dd-stage1data}

\paragraph{Data and reference annotations.}
\stageone{} uses expert segmentation masks exclusively.
The SAX pool contains 8{,}218 slices from 878 training
patients across ACDC, M\&Ms, M\&Ms-2, CMR-MULTI, and
the expert-annotated subset of In-house A.
The LAX pool contains 440 series: 200 four-chamber
series from M\&Ms-2 and 240 two- and four-chamber
series from CMR-MULTI.
The LGE pool contains 2{,}241 scar-positive slices from
MyoPS-related cohorts, EMIDEC, and CMR-MULTI.
Held-out patients are excluded before QA generation.
Annotations are mapped to a shared label set comprising
the LV cavity, LV myocardium, and RV cavity, with scar
retained for LGE; other annotation classes are excluded.

\begin{figure}[t]
    \centering
    \includegraphics[width=0.78\linewidth]{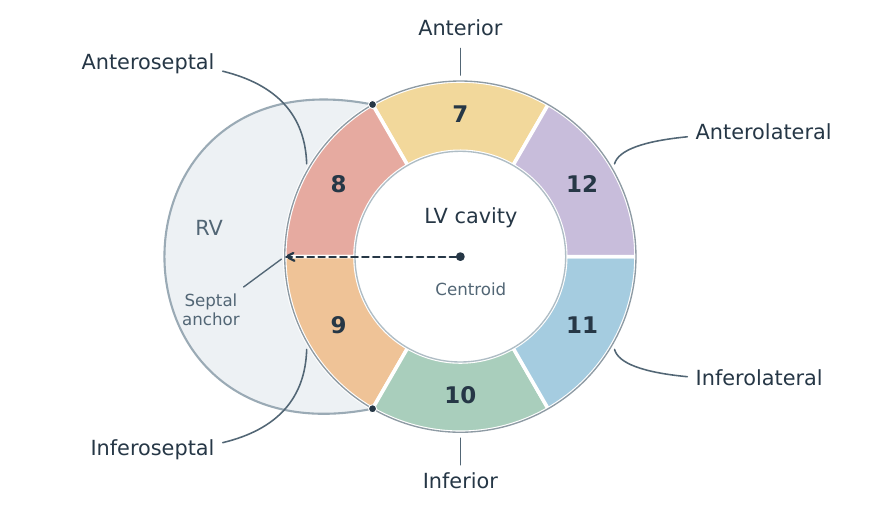}
    \caption{Mid-ventricular AHA partitioning in a canonical anatomical
    orientation. The six $60^\circ$ myocardial sectors carry AHA numbers
    7--12; these are anatomical identifiers, not the generator's local
    mask indices. The dashed ray runs from the LV-cavity centroid towards
    the septal anchor, separating the anteroseptal and inferoseptal sectors
    adjacent to the RV. Dots illustrate the two RV insertion locations.
    This is a geometric schematic, not a patient segmentation or a fixed
    scanner display orientation. Colours distinguish sectors only.}
    \label{fig:aha-mid}
\end{figure}

\paragraph{AHA-based myocardial partitioning.}
\label{app:aha-partition}
We use the 16 myocardial segments of the AHA model,
comprising six basal, six mid-ventricular, and four apical
segments, excluding the apical cap~\citep{aha17}.
For volumetric SAX data, slices containing both LV cavity
and myocardium are assigned to approximately basal,
mid-ventricular, and apical thirds, with the larger-cavity
end designated basal. Within each slice, the LV-cavity centroid defines the
polar origin. We estimate the RV insertion locations
from the myocardium adjacent to the RV and use their
septal-arc midpoint to define the septal anchor.
Acquisition orientation, where available, determines
the anterior--inferior ordering of the segments.
Basal and mid-ventricular slices are divided into six
$60^\circ$ sectors, with the anchor separating the
anteroseptal and inferoseptal segments
(Figure~\ref{fig:aha-mid}).
Apical slices use four $90^\circ$ sectors, with the
septal segment centred on the anchor.
Images, masks, and segment-label maps undergo the same
geometric transformations during augmentation,
preserving anatomical segment identities.

\begin{figure*}[t]
    \centering
    \includegraphics[width=\textwidth]{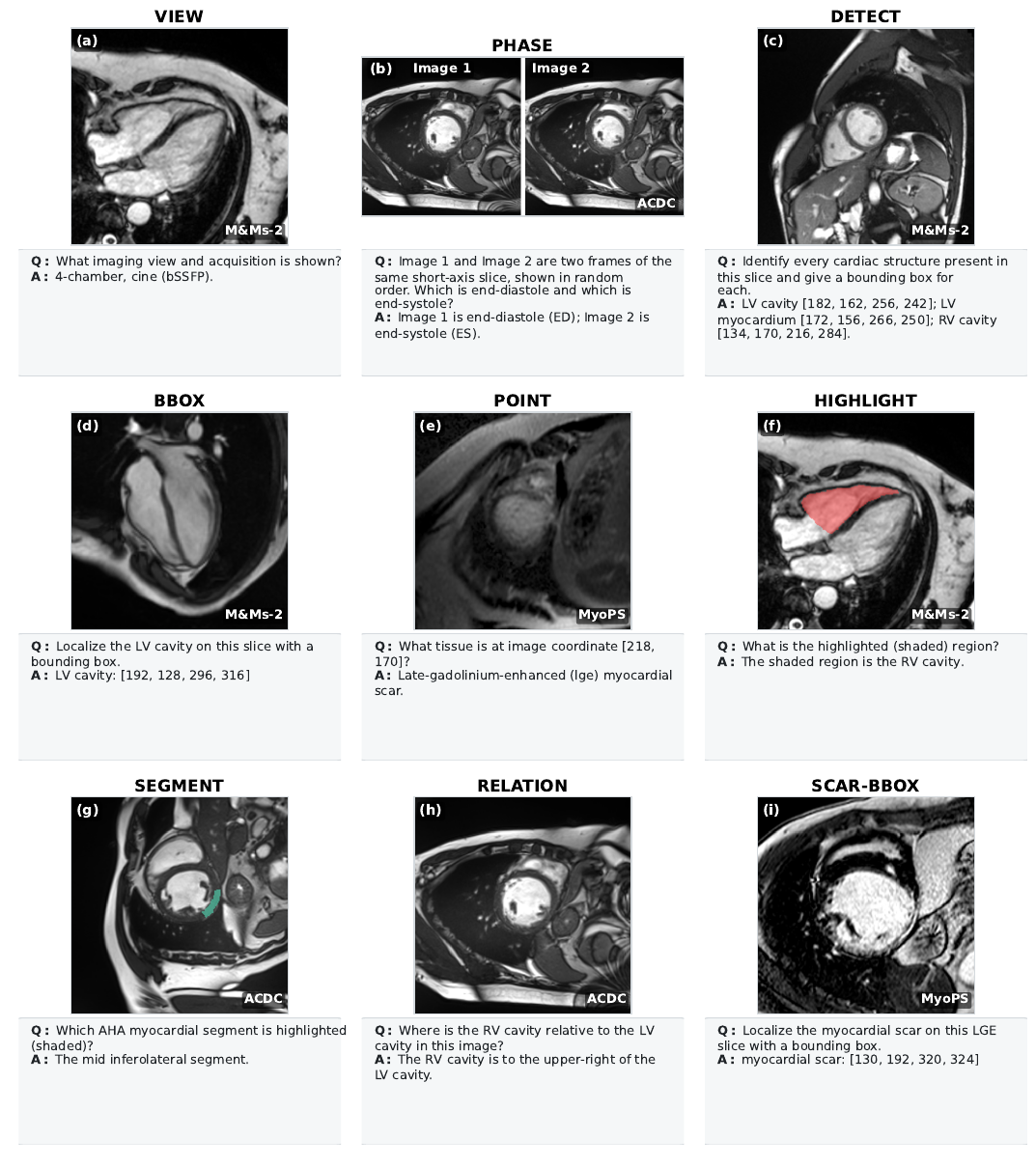}
    \caption{Examples of the nine \stageone{} grounding tasks from public
    ACDC, M\&Ms-2, and MyoPS data. Each panel shows the frozen
    $448\times448$ input, generated question, and target answer;
    \textsc{phase} uses two input frames. Coordinate targets appear only in
    text, whereas the highlights in \textsc{highlight} and \textsc{segment}
    are part of the input.}
    \label{fig:stage1-grounding-examples}
\end{figure*}

\paragraph{Grounding QA generation.}
We generate nine task types covering image context,
anatomical recognition, localisation, and spatial relations
(Table~\ref{tab:stage1tasks}).
Answers are derived from expert masks and the associated
view, sequence, and phase information.
Points and bounding boxes are represented by absolute pixel
coordinates in the $448\times448$ reference image.
Small annotated regions are filtered before localisation
questions are generated, and scar bounding boxes are
restricted to single connected components. For \textsc{highlight} and \textsc{segment}, the target region
is highlighted at 50\% opacity using one of eight colours.
\textsc{point} questions include background probes.
AHA segment questions use the labelled myocardial sectors
described in \S\ref{app:aha-partition}.
Task availability follows the annotations and imaging
context of each source.
Figure~\ref{fig:stage1-grounding-examples} shows one
representative public-data example for each task.

\paragraph{Augmentation and consistency checks.}
For cine data, horizontal and vertical flips and rotations
by $90^\circ$, $180^\circ$, and $270^\circ$ are applied
jointly to images, masks, and AHA segment maps.
SAX slices receive one sampled transformation or the identity;
LAX series retain the original images and receive three
additional sampled transformations.
The same transformation is applied to both images of a
\textsc{phase} pair, and answers are regenerated after
transformation. LGE images are not augmented.
To reduce repetitive supervision, SAX \textsc{view} and
\textsc{relation} questions are retained with probabilities
of 0.15 and 0.5, respectively.
A consistency checker finds no mismatches in 14{,}311
sampled coordinate assertions across the three domains.
This check assesses agreement between generated answers
and transformed annotations, rather than annotation accuracy.

\paragraph{Dataset composition.}
The resulting dataset contains \nQAone{} QA pairs:
77{,}482 from SAX, 35{,}040 from LAX, and 16{,}393 from LGE.
These pairs use 45{,}773 unique images, including 23{,}576
with highlight overlays; the 9{,}978 \textsc{phase} pairs
each contain two images.

\begin{table}[t]
\centering\scriptsize
\setlength{\tabcolsep}{3pt}
\caption{\stageone{} grounding tasks. Answers are generated from the ground-truth mask.}
\label{tab:stage1tasks}
\begin{tabular}{@{}lp{.50\linewidth}lr@{}}
\toprule
Task & Question $\rightarrow$ answer & Domains & QA \\
\midrule
\textsc{view} & imaging view and sequence $\rightarrow$ ``Short-axis, cine (bSSFP).'' & SAX, LAX, LGE & 7{,}005 \\
\textsc{phase} & two shuffled frames of one slice $\rightarrow$ which frame is ED and which is ES & SAX, LAX & 9{,}978 \\
\textsc{detect} & enumerate all structures $\rightarrow$ name and box for each & SAX, LAX & 11{,}738 \\
\textsc{bbox} & named structure $\rightarrow$ $[x_{\min},y_{\min},x_{\max},y_{\max}]$ & SAX, LAX & 33{,}024 \\
\textsc{point} & coordinate $[x,y]$ $\rightarrow$ structure name, or background & SAX, LAX, LGE & 35{,}796 \\
\textsc{highlight} & region highlighted in a random colour $\rightarrow$ structure name & SAX, LAX, LGE & 15{,}038 \\
\textsc{segment} & highlighted myocardial sector $\rightarrow$ AHA level and segment & SAX, LGE & 8{,}538 \\
\textsc{relation} & position of RV cavity relative to LV cavity (centroids, 8-pixel margin) & SAX, LAX & 6{,}076 \\
\textsc{scar-bbox} & single connected scar component ($\ge$15 pixels) $\rightarrow$ box & LGE & 1{,}722 \\
\midrule
Total & & & \nQAone{} \\
\bottomrule
\end{tabular}
\end{table}



\subsection{Clinical QA construction}
\label{app:dd-stage2}
\label{app:dd-labels}
\label{app:dd-formulation}

We construct clinical QA pairs from image-derived measurements
and reference annotations, using the task-specific criteria
in Table~\ref{tab:channels}. The following sections describe
reference eligibility, quality control, measurement derivation,
and training-pool assembly.

\subsubsection{Segmentation and quality control}

\paragraph{Segmentation and label harmonisation.}
Cardiac segmentation provides an anatomical basis for
quantitative assessment of cardiac structure and
function~\citep{gao2021utnet,guo2026k}.
We retain expert reference masks where available and
generate masks for the remaining images using a deployed
internal CMR segmentation pipeline.
Its native labels are mapped to a shared scheme comprising
the LV cavity, myocardium, and RV cavity, with papillary
muscle merged into the LV cavity.
Expert scar annotations are retained for LGE.
Missing annotations are treated as unavailable rather
than as evidence that a structure is absent.

\paragraph{Mask screening and QA slice selection.}
For SAX cine, masks across the full stack undergo a rapid
visual check for obvious, substantial segmentation failures,
such as missing structures or gross anatomical misalignment.
This screening does not assess fine-grained boundary accuracy.
Stack-level measurements, including EDV and EF, use the
available masks that pass screening across the full anatomical
coverage, including basal and apical levels.
Middle-level slices are selected as image inputs for SAX QA,
except for the global and regional hypertrophy tasks,
which also retain eligible basal and apical slices.
This input selection does not restrict the stack used
to derive volume-based reference labels.
QA pairs requiring both ED and ES use corresponding images
whose masks pass screening at both phases.

\paragraph{From reviewed masks to clinical QA.}
Reviewed masks provide the anatomical regions for task-specific
measurements, supplemented by acquisition metadata,
source-provided values, or tracking outputs where required.
Measurements are computed in physical units and converted
into reference answers using Table~\ref{tab:channels}.
For SAX LV-size, RV-size, and EF questions, the reference answer is
derived from the stack-level measurement and paired with
an eligible middle-level image or ED/ES image pair.
Slice-specific questions, such as hypertrophy and regional
wall motion, use measurements from the displayed slice.
An image can support multiple questions when the required
annotations and measurements are available.
The following subsections describe reference construction
for each imaging domain and clinical task.

\subsubsection{Short-axis cine reference construction}

\paragraph{Hypertrophy and regional hypertrophy.}
We estimate end-diastolic wall thickness from the LV-cavity
and myocardial masks by sampling 36 radial directions
from the LV-cavity centroid. Along each ray, the contiguous
myocardial thickness is converted to millimetres, and valid
measurements are averaged within anatomically anchored
AHA segments, using the partition described in
\S\ref{app:aha-partition}. A segment is labelled hypertrophic when
its mean thickness is at least 13\,mm. The global hypertrophy question receives a positive answer
if any segment in the displayed slice is hypertrophic.
The regional question names the affected segments or states
that none is hypertrophic. When at least five of six segments,
or three of four segments, are affected, the answer uses
the predefined description ``Concentric left-ventricular
hypertrophy in this slice.''
Both tasks use the same segment-level measurements and
therefore have consistent presence labels.

\paragraph{SAX LV size.}
We compute LV-cavity areas from the ED segmentation masks
by counting cavity pixels and multiplying by the
corresponding pixel area. These areas are summed across
the SAX stack, weighted by acquisition-specific slice
separation, to estimate LV end-diastolic volume (EDV)
in millilitres. Reference EDV values are used where available.
The absolute EDV is then mapped to normal, mild, moderate,
or severe LV dilation according to the thresholds in
Table~\ref{tab:channels}.

\paragraph{SAX ejection fraction.}
We compute LV-cavity areas from the ED and ES segmentation
masks by counting cavity pixels and multiplying by the
corresponding pixel area. Areas are summed across matched
SAX slices to estimate LV ejection fraction (EF):
\begin{equation}
\mathrm{EF}
=100\left(
1-\frac{\sum_{z\in\mathcal Z} A_z^{\mathrm{ES}}}
        {\sum_{z\in\mathcal Z} A_z^{\mathrm{ED}}}
\right),
\label{eq:stage2-ef-reference}
\end{equation}
where $A_z^{\mathrm{ED}}$ and $A_z^{\mathrm{ES}}$ are
mask-derived LV-cavity areas, and $\mathcal Z$ contains
the matched slices passing screening across the SAX stack.
For uniformly spaced slices,
the common slice separation cancels in the volume ratio.
Study-level reference EF values are used where available.
The resulting EF is classified as normal, mildly reduced,
moderately reduced, or severely reduced according to
Table~\ref{tab:channels}.

\paragraph{SAX RV size.}
We compute study-level RV EDV and the RV-to-LV EDV ratio
from ED cavity masks across the SAX stack, using the
physical-volume calculation described for LV size.
The pipeline automatically assigns a dilated label when
both RV EDV $\ge 200\,\mathrm{mL}$ and the RV-to-LV EDV
ratio $\ge 1.3$ are satisfied. Studies meeting only one
criterion are flagged for visual review of ED images and
cavity-mask overlays: confirmed cases are labelled dilated,
while unconfirmed candidates are excluded.
Studies below both thresholds form the normal sampling pool.
Each accepted study-level answer, ``RV cavity dilated'' or
``RV cavity normal'', is paired with eligible middle-level
ED/ES image pairs to generate RV-size QA.

\paragraph{Radial motion and the GRS surrogate.}
We use mean systolic wall thickening derived from ED and ES
myocardial masks as a surrogate for GRS.
For each radial sampling direction $j$, the relative
thickness change is computed as
\begin{equation}
h_j = 100\,
\frac{t_j^{\mathrm{ES}}-t_j^{\mathrm{ED}}}
     {\max(t_j^{\mathrm{ED}},1)},
\label{eq:stage2-thickening-reference}
\end{equation}
where $t_j^{\mathrm{ED}}$ and $t_j^{\mathrm{ES}}$ denote
mask-derived myocardial thicknesses in pixels.
Only directions with positive ED thickness are included,
and the denominator is bounded below by one pixel.
Relative thickness changes are averaged within six
myocardial sectors, followed by an average across valid
sectors to obtain a slice-level reference.
The resulting value is classified as normal or reduced
using the thresholds and exclusion interval in
Table~\ref{tab:channels}.

\paragraph{Regional wall motion.}
We derive regional wall-motion labels from the ED and ES
myocardial masks using the AHA 16-segment partition
described in \S\ref{app:aha-partition}.
Each middle-level SAX slice contains six AHA segments.
For each segment, we average the mask-derived radial
thickening values defined in
Equation~\ref{eq:stage2-thickening-reference}.
A segment is labelled as having reduced motion when
its mean systolic thickening is below 20\%.
The QA answer lists the affected AHA segments,
states that none has reduced motion, or indicates
that all six segments are affected.

\paragraph{Circumferential motion.}
We obtain GCS from the feature-tracking module of the
deployed CMR analysis pipeline.
A learned registration network estimates inter-frame
displacement fields, which propagate myocardial contour
points through the cine sequence.
Mid-wall circumferential strain is computed from changes
in the lengths of consecutive contour segments:
\begin{equation}
\varepsilon_{\mathrm{cc}}(t)
=\frac{L(t)-L_0}{L_0},
\end{equation}
where $L_0$ and $L(t)$ denote segment lengths in the
reference frame and at time $t$, respectively.
The module reports the peak mid-wall circumferential
strain as GCS, with more negative values indicating
greater circumferential shortening.
Cases without usable tracking results are excluded.
GCS is classified as normal or reduced using the
thresholds and exclusion interval in
Table~\ref{tab:channels}, with normal-control overrides
described below.

\subsubsection{LGE reference construction}

\paragraph{Scar presence.}
We derive scar-presence labels from expert scar masks
provided by the MyoPS-related cohorts, EMIDEC, and CMR-MULTI.
After harmonising annotation labels, an image is labelled
scar-positive if its reference scar mask is non-empty,
and scar-negative otherwise.
The label is paired with a question asking whether
myocardial scar is present in the displayed image.
Presence QA includes both SAX and LAX LGE images;
transmurality and extent QA require scar-positive SAX
images with usable myocardial and scar annotations.

\paragraph{Regional scar transmurality.}
We derive transmurality labels from expert myocardial
and scar masks. Along each radial direction, we compute
the scar extent relative to the myocardial wall thickness
and average these ratios within sliding $60^\circ$ sectors.
A slice is labelled as containing regionally transmural
scar if any sector has a mean ratio of at least 50\%;
otherwise, it is labelled non-transmural.

\paragraph{Circumferential scar extent.}
We estimate scar extent from the myocardial and scar
masks by sampling 180 radial directions from the
LV-cavity centroid.
Circumferential involvement is computed as
\begin{equation}
c =
\frac{N_{\mathrm{scar}}}{N_{\mathrm{myocardium}}},
\label{eq:scar-circumferential-extent}
\end{equation}
where $N_{\mathrm{myocardium}}$ counts rays intersecting
the myocardial wall and $N_{\mathrm{scar}}$ counts those
also intersecting scar within the wall.
This measures circumferential coverage on the displayed
slice rather than scar area or volume.
The QA answer is ``Localised scar.'' when $c<0.5$
and ``Widespread scar.'' when $c\geq0.5$.
Automatically labelled examples with $c\in[0.40,0.60)$
are excluded, while recorded manual-review decisions
can retain cases within this interval.


\subsubsection{Long-axis cine reference construction}

\paragraph{LAX ejection fraction and ventricular volume.}
LAX EF and EDV questions use two- or four-chamber
ED/ES image pairs, with the view specified in the question.
Reference labels are obtained from the same patient's
SAX grades, available reference ventricular volumes,
or measurements derived from SAX segmentation masks.
We use the same four-grade categories as the SAX tasks
(Table~\ref{tab:channels}) and pair each patient-level
label with the corresponding LAX images, maintaining
consistent reference answers across views.

\paragraph{LAX longitudinal motion.}
Training GLS references are computed from the full cine sequence
of each LAX view using the deployed feature-tracking module.
Anatomical landmarks and myocardial segmentation define contour
points, and a pretrained tracking network estimates dense
displacement fields from the first frame to subsequent frames.
Longitudinal strain at myocardial points is derived from local
deformation and myocardial orientation, then temporally smoothed.
We average strain across valid points at each frame and take the
minimum of the resulting curve:
\begin{equation}
\mathrm{GLS}=100\min_t\left(
\frac{1}{|\mathcal V_t|}\sum_{i\in\mathcal V_t}
\varepsilon_{\ell,i}(t)\right),
\end{equation}
where $\mathcal V_t$ contains the valid myocardial points at frame $t$.
Automatic myocardial masks support tracking; for M\&Ms-2,
expert ED masks replace the automatic segmentation, while motion
is still estimated from the complete image sequence. Training values are classified as normal at $\leq-16\%$ and
reduced at $\geq-12\%$, excluding intermediate, missing, and
sign-inconsistent positive values.
Each accepted label is paired with the corresponding ED/ES images
for QA input; these two frames do not limit the sequence used
for GLS computation. Eligibility is independent of EF/EDV
reference availability.

\subsubsection{QA assembly and dataset summary}

The measurements described above are converted into binary,
four-grade, or regional answers using Table~\ref{tab:channels},
subject to task-specific exclusions and recorded review corrections.
For the ACDC cohort, patients clinically diagnosed as normal
are assigned normal GRS and GCS labels before thresholding
and exclusion, with corresponding normal-reference handling
applied to hypertrophy labels. Each accepted answer is paired with a task-specific question
specifying the imaging view and anatomical context.
Hypertrophy tasks use one ED image, the remaining SAX tasks
and all LAX tasks use ED/ES pairs, and LGE tasks use one image.
The model receives images without overlays and the question;
masks support label construction and the anatomical supervision
described in Section~\ref{sec:lattn}.
The training sources in Table~\ref{tab:sources} yield
\nQAtwo{} clinical QA pairs across 14 tasks:
29{,}114 SAX, 5{,}645 LGE, and 8{,}040 LAX pairs.
Table~\ref{tab:stage2-class-distribution} summarises task-level
training counts and answer distributions.
Task counts vary with annotation availability, measurement
validity, and quality filtering.
The final RV-size training subset contains 1{,}221 QA pairs:
227 dilated and 994 normal.
The 203 non-transmural examples are oversampled to 1{,}249
records for training, producing 43{,}845 rows without adding
independent examples.

\begin{table}[!htbp]
\centering
\captionsetup{font=small,skip=4pt}
\centering
\begingroup
\CMRTableSetup
\fontsize{8}{9}\selectfont
\setlength{\tabcolsep}{3pt}
\renewcommand{\arraystretch}{1.0}
\setlength{\aboverulesep}{1pt}
\setlength{\belowrulesep}{1pt}
\caption{Clinical training QA counts before oversampling.
Four-grade counts follow the order normal, mild, moderate, and severe.}
\label{tab:stage2-class-distribution}
\begin{tabular}{@{}
p{.07\linewidth}
p{.24\linewidth}
>{\raggedleft\arraybackslash}p{.10\linewidth}
p{\dimexpr.59\linewidth-6\tabcolsep\relax}
@{}}
\toprule
Domain & Task & QA pairs & Answer counts \\
\midrule
SAX & Hypertrophy & 4{,}733 & Absent / present: 4{,}243 / 490 \\
 & Regional hypertrophy & 4{,}733 & Unaffected / affected: 4{,}243 / 490 \\
 & LV size & 3{,}985 & 2{,}963 / 412 / 304 / 306 \\
 & EF & 3{,}971 & 2{,}291 / 732 / 418 / 530 \\
 & RV size & 1{,}221 & Normal / dilated: 994 / 227 \\
 & GRS surrogate & 3{,}713 & Normal / reduced: 2{,}939 / 774 \\
 & GCS & 3{,}062 & Normal / reduced: 2{,}216 / 846 \\
 & Regional motion & 3{,}696 & Unaffected / affected: 2{,}656 / 1{,}040 \\
\midrule
LGE & Scar presence & 2{,}917 & Absent / present: 786 / 2{,}131 \\
 & Transmurality & 1{,}452 & Non-transmural / transmural: 203 / 1{,}249 \\
 & Extent & 1{,}276 & Localised / widespread: 873 / 403 \\
\midrule
LAX & EF & 2{,}725 & 1{,}752 / 493 / 189 / 291 \\
 & EDV & 2{,}747 & 2{,}011 / 304 / 242 / 190 \\
 & GLS & 2{,}568 & Normal / reduced: 2{,}125 / 443 \\
\midrule
\multicolumn{2}{@{}l}{Total} & \nQAtwo{} & \\
\bottomrule
\end{tabular}
\par\vspace{2pt}
\begin{minipage}{\linewidth}
\fontsize{7.5}{8.5}\selectfont
\textit{Note.} Regional counts group QA pairs by whether no segment
or at least one segment is affected.
\end{minipage}
\endgroup

\par\vspace{8pt}
\centering
\begingroup
\CMRTableSetup
\fontsize{8}{9}\selectfont
\setlength{\tabcolsep}{3pt}
\renewcommand{\arraystretch}{1.0}
\setlength{\aboverulesep}{1pt}
\setlength{\belowrulesep}{1pt}
\caption{Clinical test QA counts and label distributions.
Four-grade order follows Table~\ref{tab:stage2-class-distribution};
external counts use the existing model-scoring labels.}
\label{tab:test-qa-distribution}
\begin{tabular}{@{}p{.07\linewidth}p{.24\linewidth}
>{\raggedleft\arraybackslash}p{.10\linewidth}
p{\dimexpr.59\linewidth-6\tabcolsep\relax}@{}}
\toprule
\multicolumn{4}{@{}l}{\CMRGroup{A. Internal: ACDC SAX, held-out LGE, and M\&Ms-2 LAX}} \\
\midrule
Domain & Task & QA pairs & Answer counts \\
\midrule
SAX & Hypertrophy & 457 & Absent / present: 407 / 50 \\
 & Regional hypertrophy & 457 & Unaffected / affected: 406 / 51 \\
 & LV size & 171 & 116 / 21 / 22 / 12 \\
 & EF & 171 & 105 / 7 / 11 / 48 \\
 & RV size & 170 & Normal / dilated: 125 / 45 \\
 & GRS surrogate & 160 & Normal / reduced: 111 / 49 \\
 & GCS & 148 & Normal / reduced: 94 / 54 \\
 & Regional motion & 171 & Unaffected / affected: 109 / 62 \\
\midrule
LGE & Scar presence & 463 & Absent / present: 123 / 340 \\
 & Transmurality & 209 & Non-transmural / transmural: 29 / 180 \\
 & Extent & 202 & Localised / widespread: 138 / 64 \\
\midrule
LAX & EF & 160 & 93 / 32 / 13 / 22 \\
 & EDV & 160 & 115 / 19 / 19 / 7 \\
 & GLS & 121 & Normal / reduced: 87 / 34 \\
\midrule
\multicolumn{2}{@{}l}{Total} & 3{,}220 & SAX: 1{,}905; LGE: 874; LAX: 441 \\
\bottomrule
\end{tabular}
\par\vspace{4pt}
\begin{tabular*}{\linewidth}{@{\extracolsep{\fill}}lrrrrr@{}}
\toprule
\multicolumn{6}{@{}l}{\CMRGroup{B. External In-house B: same QA pool, two reference standards}} \\
\midrule
& & \multicolumn{2}{c}{Rule-based} & \multicolumn{2}{c}{Report-based} \\
\cmidrule(lr){3-4}\cmidrule(l){5-6}
Task & QA pool & Labelled QA & Negative / positive & Labelled QA & Negative / positive \\
\midrule
Hypertrophy & 1{,}945 & 1{,}945 & 1{,}646 / 299 & 1{,}936 & 1{,}359 / 577 \\
Regional hypertrophy & 1{,}831 & 1{,}831 & 1{,}646 / 185 & 1{,}822 & 1{,}357 / 465 \\
LV size & 1{,}783 & 1{,}783 & 1{,}561 / 222 & 1{,}780 & 741 / 1{,}039 \\
EF & 1{,}795 & 1{,}795 & 1{,}226 / 569 & 1{,}795 & 1{,}404 / 391 \\
RV size & 1{,}795 & 1{,}795 & 1{,}753 / 42 & 1{,}778 & 1{,}203 / 575 \\
GRS surrogate & 1{,}708 & 1{,}708 & 1{,}551 / 157 & 1{,}708 & 1{,}280 / 428 \\
GCS & 1{,}684 & 1{,}684 & 1{,}587 / 97 & 1{,}684 & 1{,}272 / 412 \\
Regional motion & 1{,}795 & 1{,}795 & 1{,}458 / 337 & 1{,}795 & 1{,}314 / 481 \\
\midrule
Total & 14{,}336 & 14{,}336 & --- & 14{,}298 & --- \\
\bottomrule
\end{tabular*}
\par\vspace{2pt}
\begin{minipage}{\linewidth}
\fontsize{7.5}{8.5}\selectfont
\textit{Notes.} Counts are per QA; external LV size/EF are binary,
and regional positives mean at least one affected region.
Report-based counts exclude 38 QA pairs without report answers.
Paired-reference counts in Table~\ref{tab:wu-label-differences}
additionally require documented, comparable fields
(Appendix~\ref{app:reference-populations}).
\end{minipage}
\endgroup

\end{table}
\clearpage

\paragraph{Clinical QA examples.}
Figures~\ref{fig:clinical-examples-sax} and~\ref{fig:clinical-examples-lge-lax}
show contrasting examples of all 14 tasks from public ACDC, MyoPS,
and M\&Ms-2 data, with unmarked input images and verbatim training QA.
Each pair shares one question; four-grade tasks show normal and severe cases.

\begingroup
\setlength{\intextsep}{8pt}
\begin{figure}[H]
    \centering
    \includegraphics[width=0.9\textwidth]{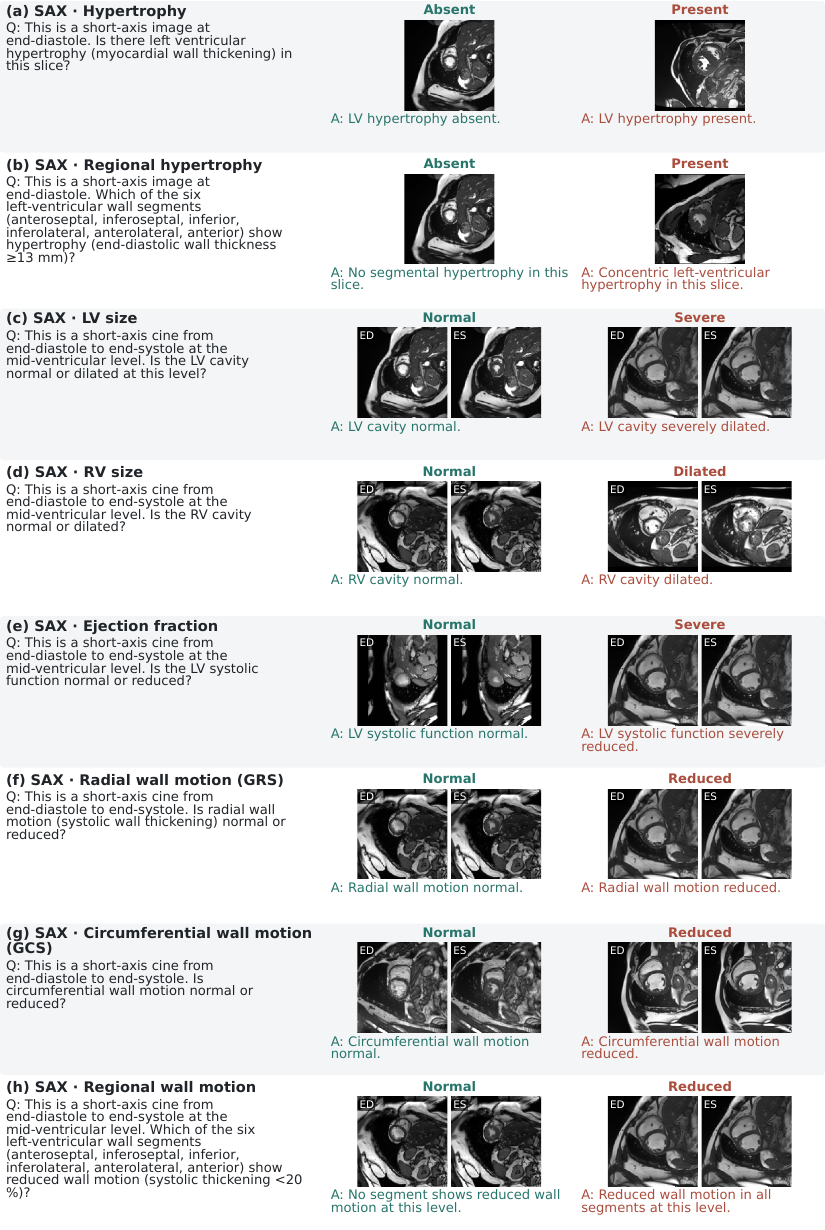}
    \caption{Clinical QA examples for the eight SAX tasks from ACDC,
    covering cardiac structure, ventricular function, and regional
    findings. Each row presents a task-specific question alongside
    two example inputs and their corresponding reference answers.}
    \label{fig:clinical-examples-sax}
\end{figure}
\begin{figure}[H]
    \centering
    \includegraphics[width=0.9\textwidth]{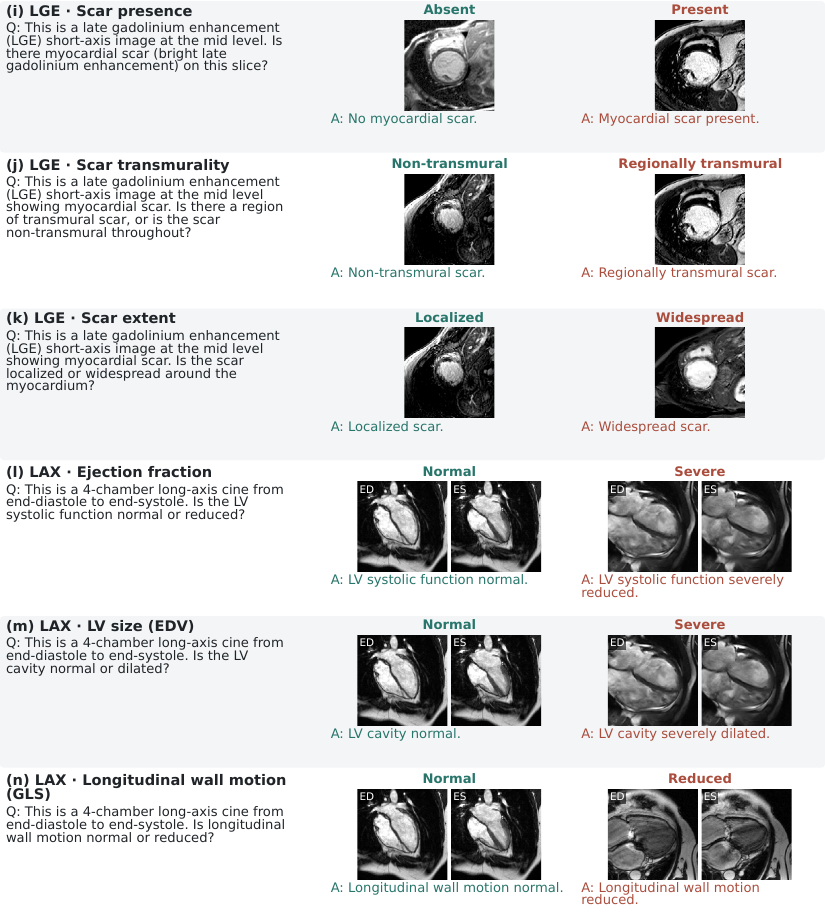}
    \caption{Clinical QA examples for the three LGE tasks from MyoPS
    and three LAX tasks from M\&Ms-2, covering scar assessment and
    ventricular function. Each row presents a task-specific question
    alongside two example inputs and their corresponding reference
    answers.}
    \label{fig:clinical-examples-lge-lax}
\end{figure}
\endgroup

\begin{table}[!t]
\centering
\caption{Reference-standard differences and paired-label disagreement in In-house B
. These are comparisons between
reference labels, not model scores. Panel B uses only documented, comparable
reference units; rates are percentages.}
\label{tab:wu-label-differences}
\begingroup
\CMRTableSetup
\fontsize{8}{9.5}\selectfont
\setlength{\tabcolsep}{3pt}
\renewcommand{\arraystretch}{1.10}
\begin{tabular}{@{}>{\raggedright\arraybackslash}p{.14\linewidth}>{\raggedright\arraybackslash}p{.28\linewidth}>{\raggedright\arraybackslash}p{.29\linewidth}>{\raggedright\arraybackslash}p{\dimexpr.29\linewidth-18pt\relax}@{}}
\toprule
\multicolumn{4}{@{}l}{\CMRGroup{A. Implemented criteria and reference scope}} \\
\midrule
Task & Rule-based reference & Report-based reference & Difference \\
\midrule
Hypertrophy & Any slice AHA wall thickness $\ge13$\,mm; reviewed basal/apical positives retained.
& Septal or posterior wall thickness $\ge11$\,mm.
& Cut-off gap: 2\,mm; slice-wide vs. two report regions. \\
\addlinespace[3pt]
LV size & Product LV EDVi: normal $\le105$; mild $(105,130]$; moderate $(130,160]$; severe $>160$.
& Report LV EDVi above its recorded upper limit: 62 (female) or 75 (male); binary dilation.
& Normal-limit gap: 43/30\,mL/m$^2$. No matched report four-grade scale. \\
\addlinespace[3pt]
RV size & Product RV EDVi $>110$\,mL/m$^2$; if BSA is absent, RV EDV $>205$\,mL.
& Explicit normal/enlarged RV description in the report conclusion.
& Quantitative rule vs. qualitative finding; not numerically comparable. \\
\addlinespace[3pt]
EF & Product EF: normal $\ge50\%$; mild $[40,50)$; moderate $[30,40)$; severe $<30\%$.
& Report EF value mapped using the same 50/40/30\% bands.
& Cut-off gap: 0 percentage points; different measurement sources. \\
\addlinespace[3pt]
GRS & Mean mask-derived systolic wall thickening: reduced $<35\%$; normal $\ge45\%$; $[35,45)$ excluded.
& Any report-mapped middle-segment WMS $\ge2$ indicates abnormal motion.
& Thickening vs. WMS proxy; not equivalent strain references. \\
\addlinespace[3pt]
GCS & Feature-tracking peak circumferential strain: normal $\le-16\%$; reduced $\ge-12\%$; $(-16,-12)$ excluded.
& Same WMS proxy as above, not measured circumferential strain.
& Strain vs. motion score; not numerically comparable. \\
\addlinespace[3pt]
Regional hypertrophy & Six middle-slice AHA segments; thickness $\ge13$\,mm.
& Two regions: septum and posterior wall; thickness $\ge11$\,mm.
& 2\,mm gap; map AS or IS to septum, IL to posterior. \\
\addlinespace[3pt]
Regional motion & Six middle-slice segments; systolic thickening $<20\%$.
& WMS $\ge2$ on five mapped segments; no anterolateral reference.
& Different quantities; compare only observed shared segments. \\
\bottomrule
\end{tabular}

\vspace{5pt}
\begin{tabular*}{\linewidth}{@{\extracolsep{\fill}}lrrrr@{}}
\toprule
\multicolumn{5}{@{}l}{\CMRGroup{B. Agreement between reference labels}} \\
\midrule
Task & \shortstack{Compared\\labels ($N$)}
& \shortstack{Rule-based\\abnormal (\%)}
& \shortstack{Report-based\\abnormal (\%)}
& \shortstack{Disagreement\\(\%)} \\
\midrule
Hypertrophy & 1,936 & 15.4 & 29.8 & 15.9 \\
LV size & 1,780 & 12.3 & 58.4 & 46.9 \\
RV size & 1,778 & 2.4 & 32.3 & 31.3 \\
EF & 1,795 & 31.7 & 21.8 & 13.3 \\
GRS vs. WMS proxy & 1,553 & 10.0 & 27.6 & 23.6 \\
GCS vs. WMS proxy & 1,528 & 6.3 & 27.0 & 24.5 \\
\addlinespace
Regional hypertrophy & 3,644 & 5.6 & 19.2 & 14.5 \\
Regional motion & 8,766 & 7.2 & 24.8 & 23.5 \\

\bottomrule
\end{tabular*}
\vspace{3pt}
\begin{minipage}{\linewidth}
\fontsize{7.5}{9}\selectfont
\textit{Notes.} $N$ counts paired QA labels for global tasks and paired
region or segment labels for regional tasks. LV size and EF are evaluated
as normal/abnormal. Disagreement is the proportion of compared labels
that differ. Missing or indeterminate reference units are excluded.
WMS denotes wall-motion score; its use for GRS/GCS provides a motion
proxy rather than a direct strain reference. Report-field availability and comparison criteria are detailed
in Appendix~\ref{app:reference-populations}.
AS/IS/IL: anteroseptal/inferoseptal/inferolateral; BSA: body surface area.
EDVi is expressed in mL/m$^2$.
\end{minipage}
\endgroup
\end{table}

\subsection{Splits and evaluation sets}
\label{app:dd-splits}

\paragraph{Patient-level partitions.}
Held-out patients are excluded across tasks, views,
and both training stages.
Internal evaluation comprises 3{,}220 QA pairs:
1{,}905 from ACDC SAX, 874 from 50 held-out LGE patients,
and 441 from M\&Ms-2 LAX.
Table~\ref{tab:test-qa-distribution} lists task-level test counts
and answer distributions, including the existing external scoring
populations under both reference standards.
Patient identities are shared across views and reconstructions
when assigning splits. In-house B is reserved exclusively
for external evaluation.

\paragraph{External cohort and dual reference labels.}
After quality filtering, In-house B contains 612 examinations
from 461 distinct patients, yielding 14{,}336 SAX QA pairs.
Repeated examinations are retained, and evaluation is performed
per QA instance.
We maintain two reference-label sources where available.
\textbf{Rule-based labels} are generated by applying task-specific
criteria to segmentation-derived measurements and other
image-analysis outputs.
\textbf{Report-based labels} are extracted from the findings,
measurements, and grades documented in the corresponding
clinical reports. QA instances without an archived task-level
report answer are excluded from evaluation. All clinical reports in In-house B follow echocardiographic
conventions for anatomical localisation and grading.
For hypertrophy and LV dilation, the adopted thresholds
are lower than the corresponding CMR thresholds, allowing
the same numerical measurement to receive different
normal/abnormal labels under the two standards.
The references also differ in anatomical coverage:
report-based regional hypertrophy covers septal and posterior
regions, while regional motion covers five mid-ventricular
segments. Report-based GRS and GCS labels use wall-motion-score
proxies rather than direct strain measurements.
We therefore evaluate the same model predictions separately
against each reference source.
Table~\ref{tab:wu-label-differences} details the implemented
criteria, reference coverage, and paired-label disagreement.
Disagreement reflects differences in measurement sources
and anatomical coverage as well as thresholds.

\paragraph{Report-derived reference availability.}
\label{app:reference-populations}
The availability of report-derived references depends on whether
relevant findings are documented and can be unambiguously extracted.
Model evaluation uses task-level report labels, whereas paired-reference
analysis additionally requires documented, comparable entries from
both sources and excludes missing or indeterminate entries.
The resulting sample counts therefore differ between
Tables~\ref{tab:test-qa-distribution} and~\ref{tab:wu-label-differences}.
Regional comparisons count matched regions or segments rather than
QA instances.

\section{Implementation details}
\label{app:training}

\subsection{Model architecture}
\label{app:cara-architecture}

\paragraph{Anatomical occupancy head.}
The shared occupancy head maps each image's
$16\times16\times2048$ post-merger features to three
occupancy maps. It comprises a $1\times1$ convolution
from 2048 to 256 channels, a $3\times3$ convolution
with 256 channels, and a $3\times3$ output convolution.
The first two layers use GELU, with dropout ($p=0.1$)
after the second. All convolutions preserve the spatial
grid. Independent sigmoid activations produce LV-cavity,
myocardial, and RV-cavity occupancies.

\paragraph{Routing and attention injection.}
Table~\ref{tab:cara-routes} specifies the anatomical prior
for each task. Each prior is peak-normalised per image
with $\epsilon=10^{-6}$ and scaled by an independent
task-specific strength $\beta_k$, initialised to 2.0
without a sign constraint.
The resulting bias is added at visual-key positions
across decoder layers, attention heads, and query positions,
preserving the original causal and padding masks
(Equation~\ref{eq:injection}).

\begin{table}[H]
\centering
\begingroup
\CMRTableSetup
\setlength{\tabcolsep}{3pt}
\caption{Anatomical priors for the 14 clinical tasks. LV and RV denote
ventricular cavities, and MYO denotes myocardium. Each task has an
independent learned injection strength.}
\label{tab:cara-routes}
\begin{tabular*}{\linewidth}{@{\extracolsep{\fill}}lll@{}}
\toprule
Domain & Clinical task & Anatomical prior \\
\midrule
SAX & Hypertrophy & MYO \\
SAX & Regional hypertrophy & MYO \\
SAX & LV size & LV \\
SAX & EF & LV \\
SAX & RV size & RV \\
SAX & Radial motion (GRS) & MYO \\
SAX & Circumferential motion (GCS) & MYO \\
SAX & Regional wall motion & MYO \\
\midrule
LGE & Scar presence & MYO \\
LGE & Scar transmurality & MYO \\
LGE & Scar extent & MYO \\
\midrule
LAX & EF & LV \\
LAX & EDV & LV \\
LAX & Longitudinal motion (GLS) & MYO \\
\bottomrule
\end{tabular*}
\endgroup
\end{table}

\subsection{Training losses}
\label{app:cara-loss}

During Stage~2, we optimise
$\mathcal L_{\mathrm{QA}}+
0.5\,\mathcal L_{\mathrm{region}}$, where
$\mathcal L_{\mathrm{QA}}$ is next-token cross-entropy
averaged over answer tokens, excluding prompt and padding
tokens, and
$\mathcal L_{\mathrm{region}}
=\mathcal L_{\mathrm{WBCE}}+\mathcal L_{\mathrm{Dice}}$.

\paragraph{Occupancy targets.}
Reference masks are average-pooled onto the visual-token
grid to obtain soft occupancy targets.
Let $p_{ir}$ and $R_{ir}$ denote the predicted and target
occupancies for cell $i$ in the local batch and structure
$r\in\{\mathrm{LV},\mathrm{MYO},\mathrm{RV}\}$.
The validity indicator $v_{ir}\in\{0,1\}$ specifies whether
the image--structure target contributes to supervision.
LGE supervision uses LV and MYO targets only.
Images without valid anatomical targets contribute
only to the QA loss.

\paragraph{Weighted binary cross-entropy.}
We use positive-class weights $w_r=(15,40,15)$ for LV, MYO,
and RV, respectively, and average over valid cell--structure pairs:
\begin{equation}
\mathcal L_{\mathrm{WBCE}}=
\frac{\sum_{i,r}v_{ir}\left[-w_rR_{ir}\log p_{ir}
 -(1-R_{ir})\log(1-p_{ir})\right]}
{\max\!\left(1,\sum_{i,r}v_{ir}\right)}.
\label{eq:region-wbce}
\end{equation}
The loss is evaluated from logits for numerical stability.

\paragraph{Soft Dice.}
Dice is computed across the local batch for each structure
and averaged over valid structures, with a smoothing constant of one:
\begin{equation}
D_r=1-
\frac{2\sum_i v_{ir}p_{ir}R_{ir}+1}
{\sum_i v_{ir}p_{ir}+\sum_i v_{ir}R_{ir}+1},
\qquad
\mathcal L_{\mathrm{Dice}}
=\frac{1}{|\mathcal R_v|}\sum_{r\in\mathcal R_v}D_r,
\label{eq:region-dice}
\end{equation}
where $\mathcal R_v=\{r:\sum_i v_{ir}>0\}$.
If no valid annotation is available, $\mathcal L_{\mathrm{region}}=0$.

Features are detached before entering $g_\phi$, so the
region loss updates the occupancy head without affecting
the visual encoder. The QA loss also updates $g_\phi$
and $\beta_k$ through the differentiable attention bias.

\subsection{Training protocol}
\label{app:training-protocol}

Stage~2 combines all 14 tasks into a shuffled training pool.
The 203 non-transmural examples are oversampled to
1{,}249 records, increasing the pool from 42{,}799 QA pairs
to 43{,}845 training records.
No other task is oversampled.
Task frequencies follow the resulting pool composition,
and all records have unit loss weight. Stage~2 uses AdamW with learning rates of $10^{-4}$ for
the LoRA parameters and occupancy head, and $10^{-2}$
for the injection strengths.
Both groups use 5\% linear warm-up followed by cosine decay.
Training uses four NVIDIA B200 GPUs with four QA instances
per GPU, giving an effective batch size of 16.

\subsection{Evaluation metrics}
\label{app:evaluation-metrics}

Scores are computed over QA instances and reported as percentages.

\paragraph{Classification.}
For class $c$, true positives ($\mathrm{TP}_c$), false positives
($\mathrm{FP}_c$), and false negatives ($\mathrm{FN}_c$) define
precision, recall, and $F_1$:
\begin{equation}
P_c=\frac{\mathrm{TP}_c}{\mathrm{TP}_c+\mathrm{FP}_c},
\qquad
R_c=\frac{\mathrm{TP}_c}{\mathrm{TP}_c+\mathrm{FN}_c},
\qquad
F_{1,c}=\frac{2\mathrm{TP}_c}
{2\mathrm{TP}_c+\mathrm{FP}_c+\mathrm{FN}_c}.
\label{eq:classification-metrics}
\end{equation}
Balanced accuracy and macro-$F_1$ average over the $C$ classes:
\begin{equation}
\mathrm{BA}=\frac{1}{C}\sum_{c=1}^{C}R_c,
\qquad
\mathrm{Macro}\text{-}F_1=\frac{1}{C}\sum_{c=1}^{C}F_{1,c}.
\label{eq:macro-metrics}
\end{equation}
Binary tasks use $C=2$; sensitivity and specificity are
positive- and negative-class recall, respectively.
PPV is positive-class precision, and ACC is the proportion
of correct predictions.
For main-table LV size, EF, and EDV, both predictions and
references are folded into normal versus abnormal before scoring.
Four-grade results use $C=4$.
SAX averages give equal weight to the six global classification
tasks, using unrounded task scores and excluding regional tasks.
External averages are computed separately for each reference.
GRS and GCS remain separate QA tasks in this average although
their Report-based labels share wall-motion-score proxies.

\paragraph{Regional localisation.}
Let $G_q$ and $\widehat G_q$ contain the reference and predicted
affected regions for question $q$, within the scored anatomical coverage.
Pooled counts and micro-$F_1$ are
\begin{equation}
\mathrm{TP}=\sum_q|G_q\cap\widehat G_q|,
\quad
\mathrm{FP}=\sum_q|\widehat G_q\setminus G_q|,
\quad
\mathrm{FN}=\sum_q|G_q\setminus\widehat G_q|,
\label{eq:regional-counts}
\end{equation}
\begin{equation}
\mathrm{Micro}\text{-}F_1=
\frac{2\mathrm{TP}}{2\mathrm{TP}+\mathrm{FP}+\mathrm{FN}}.
\label{eq:regional-micro-f1}
\end{equation}
Regional precision and recall use these pooled counts.
Exact match is the proportion of questions with
$\widehat G_q=G_q$.
Mean Jaccard averages
$|G_q\cap\widehat G_q|/|G_q\cup\widehat G_q|$ over questions,
assigning one when both sets are empty.
For both classification and regional metrics, an $F_1$
with a zero denominator is set to zero.

\paragraph{ROI attention.}
For visual token $j$, let $a_j$ be its attention weight and
$o_j\in[0,1]$ its reference occupancy for the target anatomy:
\begin{equation}
\mathrm{ROI\ attention}
=100\,\frac{\sum_{j\in\mathcal I_{\mathrm{visual}}}a_j o_j}
{\sum_{j\in\mathcal I_{\mathrm{visual}}}a_j},
\label{eq:roi-attention}
\end{equation}
where $\mathcal I_{\mathrm{visual}}$ indexes visual tokens.

\section{Additional experimental results}
\label{app:cara-additional-results}

We provide additional results for CARA-VL using the same
test sets and reference labels as the main evaluation.
Table~\ref{tab:cara-four-grade} reports four-grade
classification results for LV size, EF, and EDV.
Tables~\ref{tab:cara-internal-metrics}
and~\ref{tab:cara-external-metrics} provide detailed binary
classification metrics, with external results reported
separately against Rule-based and Report-based references.
Table~\ref{tab:cara-regional-metrics} further evaluates
regional hypertrophy and wall-motion localisation.

\begin{table}[H]
\centering
\caption{Four-grade QA classification by CARA-VL on the internal and external test sets. Grades are normal, mild, moderate, and severe. All scores are percentages.}
\label{tab:cara-four-grade}
\begingroup
\CMRTableSetup
\setlength{\tabcolsep}{3pt}
\begin{tabular*}{\linewidth}{@{\extracolsep{\fill}}llllrrrr@{}}
\toprule
Cohort & View & Reference & Task & $N$ & ACC & BA & Macro-$F_1$ \\
\midrule
ACDC-50 & SAX & Internal & LV size & 171 & 78.4 & 63.9 & 62.1 \\
ACDC-50 & SAX & Internal & EF & 171 & 79.5 & 59.3 & 53.4 \\
M\&Ms-2 & LAX & Internal & EF & 160 & 56.9 & 53.5 & 51.3 \\
M\&Ms-2 & LAX & Internal & EDV & 160 & 75.0 & 71.5 & 62.6 \\
In-house B & SAX & Rule-based & LV size & 1,783 & 70.1 & 41.4 & 31.6 \\
In-house B & SAX & Rule-based & EF & 1,795 & 52.4 & 51.6 & 44.4 \\
In-house B & SAX & Report-based & EF & 1,795 & 47.6 & 51.5 & 35.5 \\
\bottomrule
\end{tabular*}
\par\vspace{3pt}
\begin{minipage}{\linewidth}
\fontsize{8}{10}\selectfont BA is macro recall over four grades; macro-$F_1$ gives equal weight to all four classes. Report-based LV size is binary and therefore has no four-grade result. Internal SAX scores retain the archived reporting precision; the remaining scores are recomputed from the corresponding saved predictions.
\end{minipage}
\endgroup
\end{table}

\begin{table}[H]
\centering
\caption{Additional binary classification metrics for CARA-VL on the internal test sets, using the same evaluations as Table~\ref{tab:main}. All scores are percentages.}
\label{tab:cara-internal-metrics}
\begingroup
\CMRTableSetup
\setlength{\tabcolsep}{3pt}
\begin{tabular*}{\linewidth}{@{\extracolsep{\fill}}lrrrrrrr@{}}
\toprule
Task & $N$ & ACC & Sens. & Spec. & PPV & BA & Macro-$F_1$ \\
\midrule
\multicolumn{8}{@{}l}{\CMRGroup{ACDC-50 (SAX)}} \\
Hypertrophy & 457 & 93.4 & 78.0 & 95.3 & 67.2 & 86.7 & 84.3 \\
LV size & 171 & 89.5 & 90.9 & 88.8 & 79.4 & 89.9 & 88.4 \\
RV size & 170 & 91.8 & 75.6 & 97.6 & 91.9 & 86.6 & 88.8 \\
EF & 171 & 90.6 & 98.5 & 85.7 & 81.2 & 92.1 & 90.4 \\
GRS & 160 & 93.8 & 89.8 & 95.5 & 89.8 & 92.6 & 92.6 \\
GCS & 148 & 95.9 & 98.1 & 94.7 & 91.4 & 96.4 & 95.7 \\
\addlinespace[4pt]
\multicolumn{8}{@{}l}{\CMRGroup{LGE test (LGE)}} \\
Scar presence & 463 & 85.5 & 88.5 & 77.2 & 91.5 & 82.9 & 82.0 \\
Scar transmurality & 209 & 88.0 & 95.6 & 41.4 & 91.0 & 68.5 & 71.1 \\
Scar extent & 202 & 84.7 & 76.6 & 88.4 & 75.4 & 82.5 & 82.3 \\
\addlinespace[4pt]
\multicolumn{8}{@{}l}{\CMRGroup{M\&Ms-2 (LAX)}} \\
EF & 160 & 75.0 & 97.0 & 59.1 & 63.1 & 78.1 & 74.9 \\
EDV & 160 & 83.1 & 88.9 & 80.9 & 64.5 & 84.9 & 81.0 \\
GLS & 121 & 85.1 & 58.8 & 95.4 & 83.3 & 77.1 & 79.6 \\
\bottomrule
\end{tabular*}
\par\vspace{3pt}
\begin{minipage}{\linewidth}
\fontsize{8}{10}\selectfont ACC denotes accuracy; Sens. and Spec. denote sensitivity and specificity; PPV denotes positive predictive value (precision). LV size, EF, and EDV are folded to normal versus any abnormal grade. Classification $F_1$ is macro-averaged over both classes. Positive labels are hypertrophy, cavity dilation, reduced function/motion, scar presence, transmural scar, and widespread scar, respectively.
\end{minipage}
\endgroup
\end{table}

\begin{table}[H]
\centering
\caption{Additional binary classification metrics for CARA-VL on In-house B under the two reference standards. All scores are percentages.}
\label{tab:cara-external-metrics}
\begingroup
\CMRTableSetup
\setlength{\tabcolsep}{3pt}
\begin{tabular*}{\linewidth}{@{\extracolsep{\fill}}lrrrrrrr@{}}
\toprule
Task & $N$ & ACC & Sens. & Spec. & PPV & BA & Macro-$F_1$ \\
\midrule
\multicolumn{8}{@{}l}{\CMRGroup{Rule-based}} \\
Hypertrophy & 1,945 & 95.6 & 74.9 & 99.3 & 95.3 & 87.1 & 90.7 \\
LV size & 1,783 & 76.1 & 91.0 & 74.0 & 33.2 & 82.5 & 66.6 \\
RV size & 1,795 & 94.9 & 28.6 & 96.5 & 16.4 & 62.5 & 59.1 \\
EF & 1,795 & 66.5 & 92.6 & 54.4 & 48.5 & 73.5 & 66.3 \\
GRS & 1,708 & 95.1 & 77.7 & 96.9 & 71.8 & 87.3 & 86.0 \\
GCS & 1,684 & 85.6 & 82.5 & 85.8 & 26.2 & 84.1 & 65.8 \\
\addlinespace[4pt]
\multicolumn{8}{@{}l}{\CMRGroup{Report-based}} \\
Hypertrophy & 1,936 & 81.9 & 40.0 & 99.7 & 98.3 & 69.9 & 72.7 \\
LV size & 1,780 & 60.9 & 45.6 & 82.3 & 78.3 & 64.0 & 60.7 \\
RV size & 1,778 & 68.4 & 7.3 & 97.6 & 59.2 & 52.4 & 46.8 \\
EF & 1,795 & 59.5 & 95.9 & 49.4 & 34.5 & 72.6 & 58.2 \\
GRS$^{*}$ & 1,708 & 79.5 & 29.0 & 96.4 & 72.9 & 62.7 & 64.5 \\
GCS$^{*}$ & 1,684 & 84.7 & 55.8 & 94.1 & 75.4 & 75.0 & 77.2 \\
\bottomrule
\end{tabular*}
\par\vspace{3pt}
\begin{minipage}{\linewidth}
\fontsize{8}{10}\selectfont ACC denotes accuracy; Sens. and Spec. denote sensitivity and specificity; PPV denotes positive predictive value (precision). LV size, EF, and EDV are folded to normal versus any abnormal grade. Classification $F_1$ is macro-averaged over both classes. The two reference blocks score the same predictions using the task-specific coverage of Table~\ref{tab:wu-sax-qa}. $^{*}$Report GRS/GCS labels are wall-motion-score proxies, not measured strain. The paired-reference analysis additionally excludes missing or indeterminate reference fields (Appendix~\ref{app:reference-populations}).
\end{minipage}
\endgroup
\end{table}

\begin{table}[H]
\centering
\caption{Additional regional-task metrics for CARA-VL. Precision, recall, and micro-$F_1$ are pooled over question--region/segment pairs; exact match and mean Jaccard compare the affected-region sets per QA. All scores are percentages.}
\label{tab:cara-regional-metrics}
\begingroup
\CMRTableSetup
\setlength{\tabcolsep}{3pt}
\begin{tabular*}{\linewidth}{@{\extracolsep{\fill}}lllrrrrrr@{}}
\toprule
Cohort & Reference & Task & $N$ & Prec. & Rec. & Micro-$F_1$ & Exact & Jaccard \\
\midrule
ACDC-50 & Internal & Hypertrophy & 457 & 65.8 & 58.3 & 61.8 & 88.2 & 91.3 \\
ACDC-50 & Internal & Wall motion & 171 & 51.1 & 61.5 & 55.8 & 57.9 & 71.1 \\
In-house B & Rule-based & Hypertrophy & 1,831 & 71.6 & 44.8 & 55.1 & 91.2 & 93.3 \\
In-house B & Rule-based & Wall motion & 1,795 & 33.9 & 40.1 & 36.7 & 76.7 & 80.2 \\
In-house B & Report-based & Hypertrophy & 1,822 & 94.9 & 18.5 & 30.9 & 74.8 & 78.0 \\
In-house B & Report-based & Wall motion & 1,795 & 69.4 & 22.9 & 34.5 & 69.4 & 74.6 \\
\bottomrule
\end{tabular*}
\par\vspace{3pt}
\begin{minipage}{\linewidth}
\fontsize{8}{10}\selectfont $N$ counts QA instances. Report-based hypertrophy covers septal and posterior regions; report-based wall motion covers the five mapped middle segments. Exact match requires equality of the predicted and reference affected-region sets within the scored coverage. Jaccard is one when both sets are empty. These are regional localisation metrics, not pixel-level segmentation scores. No regional metric is included in classification averages.
\end{minipage}
\endgroup
\end{table}

\end{document}